\documentclass[lettersize,journal]{IEEEtran}
\usepackage{subfigure}
\usepackage{amsmath,amsfonts}
\usepackage[linesnumbered,ruled,vlined]{algorithm2e}
\usepackage{array}
\usepackage{textcomp}
\usepackage{stfloats}
\usepackage{url}
\usepackage{verbatim}
\usepackage{graphicx}
\usepackage{cite}

\usepackage{color}
\usepackage{booktabs}
\usepackage{threeparttable}
\usepackage{bbding}
\usepackage{pifont}
\usepackage{diagbox}
\usepackage{hyperref}
\usepackage{enumitem}
\usepackage{tabularx}
\usepackage{multicol}
\usepackage{xcolor}

\usepackage{amssymb}
\usepackage{amsthm}
\newtheorem{theorem}{Theorem}

\begin{document}

\title{STeP-Diff: Spatio-Temporal Physics-Informed Diffusion Models for Mobile Fine-Grained Pollution Forecasting
}

\author{
Nan Zhou, 
Weijie Hong,
Huandong Wang, ~\IEEEmembership{Member, ~IEEE},
Jianfeng Zheng,
Qiuhua Wang,
Yali Song,
Xiao-Ping Zhang, ~\IEEEmembership{Fellow, ~IEEE},
Yong Li, ~\IEEEmembership{Senior Member, ~IEEE},
and Xinlei Chen, ~\IEEEmembership{Member, ~IEEE}

\thanks{Xinlei Chen, Xiao-Ping Zhang, Nan Zhou are with Shenzhen International Graduate School, Tsinghua University, China (e-mail: \{chen.xinlei, xiaoping.zhang\}@sz.tsinghua.edu.cn;  zhoun24@mails.tsinghua.edu.cn).}

\thanks{Yong Li, Huandong Wang are with the Department of Electronic Engineering, Beijing National Research Center for Information Science and Tech- nology (BNRist), Tsinghua University, China
(e-mail: \{liyong07, wanghuandong\}@tsinghua.edu.cn).}

\thanks{Weijie Hong and Jianfeng Zheng are with the Shenzhen Smartcity Communication, Shenzhen, China (e-mail:  \{hongweijie, zhengjianfeng\}@smartcitysz.com).}

\thanks{Yali Song is with College of Soil and Water Conservation, Southwest Forestry University, China.
(e-mail: songyali@swfu.edu.cn).}
\thanks{Qiuhua Wang is with College of Civil Engineering, Southwest Forestry University, China.
(e-mail: qhwang2010@swfu.edu.cn).}

\thanks{Corresponding author: Xinlei Chen, Huandong Wang.}

\thanks{The code can be accessed at \href{https://github.com/Munan222/STeP-Diff-for-Air-Pollution-Forecasting}{https://github.com/Munan222/STeP-Diff-for-Air-Pollution-Forecasting}.}
}


\markboth{IEEE Transactions on Knowledge and Data Engineering}%
{Zhou. N \MakeLowercase{\textit{et al.}}: STeP-Diff}

\IEEEpubid{0000--0000/00\$00.00~\copyright~2021 IEEE}

\maketitle

\begin{abstract}
% Fine-grained air pollution forecasting is crucial for urban management and the development of healthy buildings. Compared to expensive and limited static monitoring stations, deploying portable sensors on mobile platforms such as cars and buses offers a low-cost and wide-coverage data collection solution. However, the random and uncontrollable movement patterns of these non-dedicated mobile platforms result in sensor data coverage that is highly sparse and time-varying. 
Fine-grained air pollution forecasting is crucial for urban management and the development of healthy buildings. Deploying portable sensors on mobile platforms such as cars and buses offers a low-cost, easy-to-maintain, and wide-coverage data collection solution. However, due to the random and uncontrollable movement patterns of these non-dedicated mobile platforms, the resulting sensor data are often incomplete and temporally inconsistent.
By exploring potential training patterns in the reverse process of diffusion models, we propose \underline{S}patio-\underline{Te}mporal \underline{P}hysics-Informed \underline{Diff}usion Models (STeP-Diff). 
STeP-Diff leverages DeepONet to model the spatial sequence of measurements along with a PDE-informed diffusion model to forecast the spatio-temporal field from incomplete and time-varying data. Through a PDE-constrained regularization framework, the denoising process asymptotically
% {\color{blue}converges} 
converges
to the convection–diffusion dynamics, ensuring that predictions are both grounded in real-world measurements and aligned with the fundamental physics governing pollution dispersion. 
To assess the performance of the system, we deployed 59 self-designed portable sensing devices in two cities, operating for 14 days to collect air pollution data. 
Compared to the second-best performing algorithm, our model achieved improvements of up to 89.12\% in MAE, 82.30\% in RMSE, and 25.00\% in MAPE, with extensive evaluations demonstrating that STeP-Diff effectively captures the spatio-temporal dependencies in air pollution fields.
% Compared to the second-best performing algorithm, our model achieved improvements of up to 89.12\%, 82.30\%, and 25.00\% in MAE, RMSE, and MAPE, respectively.
% Extensive evaluations demonstrate that STeP-Diff effectively captures the spatio-temporal dependencies in air pollution fields, achieving superior prediction accuracy.
\end{abstract}

\begin{IEEEkeywords}
Air Pollution, Mobile Sensing, Spatio-Temporal Forecasting, Diffusion Model.
\end{IEEEkeywords}

% \IEEEpeerreviewmaketitle

% \vspace{-0.4em}
% \IEEEraisesectionheading{\section{Introduction}}
\section{Introduction}

\IEEEPARstart{E}{ffective} monitoring of air pollution is crucial for safeguarding public health. 
According to the World Health Organization, approximately 7 million premature deaths each year are attributed to diseases exacerbated by air pollution \cite{liu2024mobiair, 10015790, rentschler2023global}.
Currently, urban air quality monitoring primarily relies on fixed national stations. However, these stations are limited in number and fixed in location, providing only a coarse overview of urban air quality. This limitation is especially concerning, given that major air pollutants, such as PM$_{2.5}$, NO$_2$, and CO, exhibit significant concentration gradients even over relatively short distances (on the order of hundreds of meters) \cite{10.1145/3517227}. 
To improve the granularity of air quality monitoring, mobile crowd-sensing technology has shown great potential \cite{10.1145/3397328}. This technology leverages sensor-equipped mobile platforms (e.g., cars, buses, drones), which continuously collect and transmit air pollution concentration data and their sampling locations during travel, providing extensive coverage and relatively fine-grained spatio-temporal measurements \cite{wang2025ultra, chen2017design}. 
% \cite{ma2020fine}.

% When estimating fine-grained air pollution using sensor data, two main types of models are employed: data-driven models and physics-based models. 
% Data-driven models rely on environmental data, employing methods such as graph neural networks, or generative models to learn the underlying patterns within the data
% \cite{9309378, han2022semi, qi2018deep}. In recent years, diffusion models have made significant progress in time-series inference tasks, thanks to their ability to gradually transform noise into plausible samples \cite{Tashiro2021}. 
% These models excel with abundant sensor data, effectively capturing complex patterns and dynamics. However, their performance degrades significantly with incomplete or insufficient data.
% In contrast, physics-based models construct parameterized and deterministic descriptions based on physical laws and domain knowledge, aiming to model the diffusion process of air pollution over time and space \cite{Chen2020, shi2025phy}. 
% However, these models have limited fitting capacity and struggle to accurately capture the complex interactions among high-dimensional variables. 

% {\color{blue}
Fine-grained air pollution estimation using sensor data typically falls into three categories: physics-based, data-driven, and hybrid physics–data-driven methods.
Physics-based methods simulate pollutant diffusion using physical laws and domain knowledge \cite{Holmes06, zannetti2013air}, offering strong interpretability but a limited ability to capture complex high-dimensional interactions.
Data-driven methods use models like GNNs or diffusion models to learn latent patterns from environmental data \cite{9309378, han2022semi, qi2018deep, Tashiro2021}, performing well with dense data but struggling with sparsity or incompleteness.
Hybrid methods integrate physical constraints into data-driven models via loss regularization or architectural design \cite{hettige2024airphynet,shi2025phy,tian2024air}. 
However, neural representations are often nonlinear abstractions, making them hard to align with explicit physical variables—posing a key challenge in model integration.
% }

\IEEEpubidadjcol
 
Mobile platforms mitigate the high costs and limited coverage of fixed air pollution monitoring stations; however, their uncontrolled and erratic movement patterns result in incomplete and time-varying data coverage.
Therefore, the core problem addressed in this paper is: \textit{How can we leverage incomplete and time-varying data collected from non-dedicated mobile platforms to forecast fine-grained spatio-temporal pollution fields? } 
During this process, we face two major challenges:
\textit{(C1) Incomplete and Time-Varying Data Coverage}: Due to the concentration of most mobile sensing nodes in urban centers or busy areas, coupled with their non-stationary trajectories, the time-series data at individual locations often exhibit irregularities, thereby complicating the accurate inference of spatio-temporal pollution patterns.
\textit{(C2) Inaccurate Physical Knowledge}: 
The air pollution diffusion process is complex and large-scale, with existing domain knowledge often being incomplete, making accurate modeling challenging.
In summary, traditional data-driven models struggle with incomplete and time-varying data, while purely physics-based models are hindered by inadequate domain knowledge and approximation errors, preventing them from fully capturing real-world pollution diffusion.

To overcome these challenges, we propose Spatio-Temporal Physics-Informed Diffusion Models (STeP-Diff), which directly learn the conditional distribution of air pollution fields. Specifically:
\textit{(S1) To address (C1)}: 
Building upon the unique capability of diffusion models to learn data distributions through iterative denoising, we innovatively incorporate latent spatial features and historical observations as conditional inputs into the reverse process. This enhances the model’s ability to capture dynamic pollution fields, even under incomplete and time-varying data coverage.
\textit{(S2) To address (C2)}: 
% We embed the limited PDE-based physical prior knowledge into the loss function as a regularization constraint to ensure that the denoising process progressively approximates convection-diffusion dynamics. Furthermore, by leveraging the unique capability of DeepONet to capture latent spatial patterns, we inject essential spatial information into the diffusion model. 
% This approach efficiently integrates physical knowledge into the learning process, addressing the issues of inaccurate physical knowledge.
% We embed limited PDE-based physical priors into the loss function as a regularization constraint to ensure that the denoising process progressively approximates convection–diffusion dynamics. Furthermore, by leveraging the unique capacity of DeepONet to capture latent spatial patterns, we inject essential spatial information into the diffusion model. 
% This approach effectively integrates physical knowledge into the learning process, mitigating the impact of incomplete or inaccurate domain knowledge.
We embed PDE-based physical priors into the loss function as a regularization constraint to ensure that the denoising process approximates convection–diffusion dynamics. Furthermore, by leveraging the capacity of DeepONet to capture latent spatial patterns, we inject essential spatial information into the diffusion model. This approach effectively integrates physical knowledge into the learning process, mitigating the impact of incomplete or inaccurate domain knowledge.

% The main contributions of this paper are summarized as follows:
The main contributions of this paper are as follows:
\begin{itemize}
    \item We introduce STeP-Diff and successfully apply it to mobile fine-grained air quality forecasting.
    Additionally, we design an innovative self-supervised training strategy to optimize the conditional diffusion model.
    \item 
    By embedding customized conditional distributions and PDE constraints, the denoising process is guided not only by the observational data but also by the inherent diffusion dynamics, thereby imbuing the learning procedure with physical interpretability while enhancing accuracy.
    \item 
    % We deployed 59 self-designed portable sensing devices across two cities over a 14-day period to collect air pollution data and evaluate model performance. 
    % Extensive evaluation results highlight the exceptional predictive capability of our model in spatio-temporal air pollution field forecasting.
    We deployed 59 self-designed portable sensors in two cities over 14 days to collect air pollution data and evaluate model performance. Extensive results demonstrate the exceptional predictive capability of our model in spatio-temporal pollution field forecasting.
\end{itemize}

% The remainder of the paper is structured as follows: Section \ref{pre} introduces the mathematical model and provides an overview of the system; Section \ref{met} presents the spatio-temporal pollution field forecasting method based on the diffusion model; Section \ref{eval} details the design and deployment of the sensing devices, the analysis of the collected measurements, and the comparison of algorithm performance; Section \ref{rel} reviews related work and discusses its applications; and Section \ref{con} summarizes our work and presents future directions.
The remainder of the paper is structured as follows: Section \ref{pre} introduces the mathematical model and overviews the system; Section \ref{met} presents the spatio-temporal pollution field forecasting method based on the diffusion model; Section \ref{eval} details the design and deployment of the sensing devices, the analysis of the collected measurements, and the comparison of algorithm performance; Section \ref{rel} reviews related work and discusses its applications; and Section \ref{con} summarizes our work and presents future directions.
\section{Preliminaries and system overview}
\label{pre}

In this section, we present the mathematical setup in Section~\ref{math}, describe the problem in Section~\ref{2.2} and provide an overview of our system in Section~\ref{2.3}. For the sake of clarity, we summarize the key symbols used throughout the paper in Table~\ref{notations}.

\begin{table}
  \caption{Notations and descriptions.}
  \label{notations}
  \normalsize
  \begin{tabular}{cl}
    \toprule
    Notation & Description\\
    \midrule
    $V$ & Incomplete spatio-temporal pollution measurement \\
    $V^\text{co}$ & Observed pollution measurement \\
    $V^\text{ta}$ & Forecasting targets measurement \\
    $V^\text{de}$ & The prediction output of DeepONet\\
    $V^\text{info}$ & The complete set of conditional information \\ & provided to the Diffusion model\\
    $V^{\tau}$ & Observed pollution measurement at time $\tau$\\
    $V^{l}$ & Observed pollution measurement at discrete time
    \\ & slice $l$, with a time interval of $\Delta\tau$ \\
    $M$ & Incomplete spatio-temporal measurement mask\\
    $M^\text{co}$ & Observed measurement mask\\
    $M^\text{ta}$ & Forecasting targets mask\\
    $L$ & Time silces\\
    $L^\text{co}$ & Observation period\\
    $L^\text{ta}$ & Forecasting targets period\\
    $\tau$ & Time step\\
    $t$ & Diffusion step\\
    $q(\cdot)$ & The true data distribution\\
    $p_\theta(\cdot)$ & The data distribution learned by the model\\
    $\mu_\theta$ & Mean of $p_\theta(\cdot)$ \\
    $\sigma_\theta$ & Variance of $p_\theta(\cdot)$ \\
    $\epsilon_\theta$ & Trainable denoising function \\
    $F(\cdot)(\cdot)$ & DeepONet operator \\
    $G(\cdot)$ & PDE operator\\
    $K$ & Diffusion coefficient\\
    $P$ & Wind velocity vector\\
    $R$ & Environment factor\\
  \bottomrule
\end{tabular}
\end{table}

\vspace{-0.2cm}
\subsection{Mathematical Setup}
\label{math}

The objective of our system is to find the optimal predictions of air pollution values for each subregion. To frame air pollution estimation as an optimization problem, we consider spatio-temporal sequences with missing values in the $[X, Y]$ latitude-longitude coordinate space. The spatial domain is discretized into square subregions of size $\Delta n \times \Delta n$, and the temporal domain is discretized into time slices of length $\Delta l$. $\Delta n$ and $\Delta l$ correspond to the spatial and temporal resolutions, respectively. 
% {\color{blue}
Based on existing literatures \cite{Chen2020,qi2018deep}  and the spatiotemporal density characteristics of our dataset, we set $\Delta n = 500$m and $\Delta l = 1$ hour in our experiments.
% }
For computational convenience, we represent the observed measurements of each spatio-temporal sequence as $V = \{v^{1:L, 1:X, 1:Y}\} \in \mathcal{R}^{L \times X \times Y}$, where $L$ is the number of time slices, $X$ is the number of longitudinal subregions, and $Y$ is the number of latitudinal subregions. While the length $L$ of each time series may vary, for simplicity, unless otherwise stated, we assume that the length of all time series is the same. We also represent the observation mask as $M = \{m^{1:L, 1:X, 1:Y}\} \in \{0, 1\}^{L \times X \times Y}$, where if $v^{l, x, y}$ is missing, then $m^{l, x, y} = 0$; if $v^{l, x, y}$ is observed, then $m^{l, x, y} = 1$. Thus, each spatio-temporal sequence is represented as $\{V, M\}$.

\subsection{Problem Description}
\label{2.2}

Based on the aforementioned mathematical definitions, the problem under investigation can be formulated as follows:

% \textit{Spatio-Temporal Air Pollution Forecasting Modeling}:

% \textit{Given}: The observed sparse spatio-temporal pollution fields, $\{V^{co},M^{co}\}$, during the observation period  $L^{co}$, and the forecasting target period $L^{ta}$.

% \textit{Problem}: Predict accurate and robust fine-grained pollution measurements, $V^{ta}$, for the target time period $L^{ta}$.

\textit{Definition 1 (Spatio-Temporal Air Pollution Forecasting Modeling)}.
Given observed incomplete spatio-temporal pollution data, denoted as $\{V^\text{co}, M^\text{co}\}$, over an observation period $L^\text{co}$, and a target forecasting period $L^\text{ta}$, the objective is to predict accurate and robust fine-grained pollution measurements, $V^\text{ta}$, for the target period $L^\text{ta}$.

\vspace{-0.2cm}
\subsection{System Overview}
\label{2.3}

\begin{figure*}[t]
  \centering
  \includegraphics[width=0.95\linewidth]{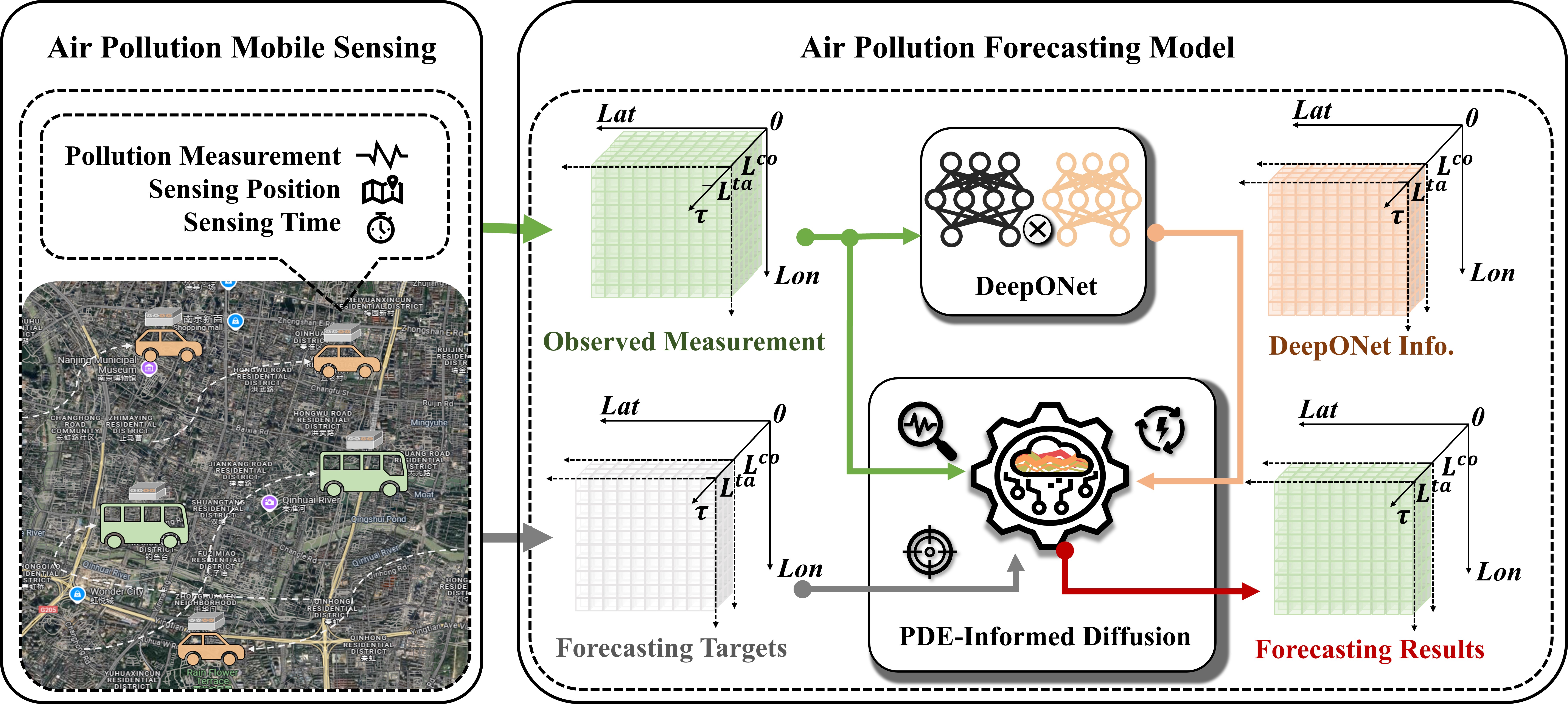}
  \caption{Our air pollution forecasting framework, STeP-Diff, consists of two primary components: Air Pollution Mobile Sensing and the Air Pollution Forecasting Model. The Mobile Sensing component transmits observed measurement data and forecast targets to the Forecasting Model. The Forecasting Model then leverages two core components—DeepONet and PDE-informed Diffusion—to generate accurate predictions.}
  \label{overview}
\end{figure*}

The primary objective in developing STeP-Diff is to minimize the prediction error in air pollution levels. As illustrated in Fig. \ref{overview}, the system architecture of STeP-Diff comprises two major modules: Air Pollution Mobile Sensing and Air Pollution Forecasting Model. 

Initially, air quality data and corresponding vehicle information are collected using mobile sensors and speedometers. To enhance computational efficiency, the Mobile Sensing module discretizes both the temporal and spatial domains into uniform subregions and time slices, assigning data collected from various mobile nodes to the corresponding subregion and time slice. When multiple data points are acquired within the same subregion and time slice, their average value is adopted as the representative measurement for that spatio-temporal unit.

Subsequently, the processed historical sensor measurements and prediction targets are fed into the Forecasting Model. First, DeepONet is employed to extract pollution features at various spatial points. These features, along with historical observations and prediction targets, are then aligned and input into the PDE-informed diffusion model for prediction. By incorporating physical constraints, the model not only relies on observed measurements but also gradually adheres to the molecular dynamics of pollutant diffusion, ensuring that the predictions possess theoretical interpretability. 
% We innovatively embed the underlying physical principles into the learning process to address the task of fine-grained, city-scale spatio-temporal pollution concentration field forecasting, providing a solid technical foundation for air quality prediction and pollution control.

% \section{\color{blue}Methodology}
\section{Methodology}

\label{met}

In this section, we present our Air Pollution Forecasting Model. We begin by briefly reviewing Denoising Diffusion Probabilistic Models (DDPM) and introduce how predictions can be made based on DDPM in Section \ref{3.1}. Next, we propose 
% the Spatio-Temporal Physics-Informed Diffusion Models (STeP-Diff)
STeP-Diff
in Section \ref{3.2}, which consists of two core components: DeepONet and the PDE-Informed Diffusion Model. Finally, we provide a detailed description of the algorithm implementation for spatio-temporal field forecasting using the STeP-Diff in Section \ref{3.3}.

\subsection{Forecasting with Diffusion Models}
\label{3.1}
\subsubsection{Denoising Diffusion Probabilistic Models (DDPM)}
\label{DDPM}
We consider learning a model distribution \( p_\theta(v_0) \), which approximates the data distribution \( q(v_0) \). Let \( v_t \), for \( t = 1, \dots, T \), represent a sequence of latent variables in the same sample space as \( v_0 \), denoted as $ \mathcal{V} $. DDPM \cite{Ho2020} is a latent variable model consisting of two processes: the forward process and the reverse process. The forward process is defined by the following Markov chain:
\begin{equation}
\begin{aligned}
& q(v_{1:T} \mid v_0) := \prod_{t=1}^{T} q(v_t \mid v_{t-1}),  \\
& q(v_t \mid v_{t-1}) := \mathcal{N}\left(\sqrt{1-\beta_t} v_{t-1}, \beta_t I\right), \\
& q(v_t \mid v_0) = \mathcal{N}(v_t; \sqrt{\alpha_t} v_0, (1-\alpha_t) I),  \alpha_t := \prod_{i=1}^{t} \hat{\alpha}_i, \hat{\alpha}_t := 1 - \beta_t, 
\end{aligned}
\end{equation}
where $\alpha_t, \beta_t $ are small positive constants representing the noise level. Consequently, \( v_t \) can be expressed as:
\begin{equation}
v_t = \sqrt{\alpha_t} v_0 + (1 - \alpha_t) \epsilon, \epsilon=\sum_{j=1}^t {\epsilon_j}, \epsilon_j \sim \mathcal{N}(0, I).
\label{noise-added}
\end{equation}

In contrast, the reverse process aims to denoise \( v_t \) to recover \( v_0 \), and is defined by the following Markov chain:
\begin{equation}
\begin{aligned}
& p_{\theta}(v_{0:T}) := p(v_T) \prod_{t=1}^{T} p_{\theta}(v_{t-1} \mid v_t), v_T \sim \mathcal{N}(0, I),\\
& p_{\theta}(v_{t-1} \mid v_t) := \mathcal{N}(v_{t-1}; \mu_{\theta}(v_t, t), \sigma_{\theta}(v_t, t) I), 
\end{aligned}
\label{reverse}
\end{equation}
where $\mu_{\theta}(v_t, t) = \frac{1}{\alpha_t} \left( v_t - \frac{\beta_t}{\sqrt{1 - \alpha_t}} \epsilon_{\theta}(v_t, t) \right),
\sigma_{\theta}(v_t, t) = \hat{\beta}_t^{1/2},  \hat{\beta}_t = 
\begin{cases} 
\frac{1 - \alpha_{t-1}}{1 - \alpha_t} \beta_t & t > 1 \\
\beta_1 & t = 1 
\end{cases} $,
% \begin{equation*}
% \text{where }
% \mu_{\theta}(v_t, t) = \frac{1}{\alpha_t} \left( v_t - \frac{\beta_t}{\sqrt{1 - \alpha_t}} \epsilon_{\theta}(v_t, t) \right), \quad
% \sigma_{\theta}(v_t, t) = \hat{\beta}_t^{1/2}, \quad \hat{\beta}_t = 
% \begin{cases} 
% \frac{1 - \alpha_{t-1}}{1 - \alpha_t} \beta_t & t > 1 \\
% \beta_1 & t = 1 
% \end{cases} ,
% \end{equation*}
and \( \epsilon_\theta \) is the trainable denoising function. 
% The denoising function corresponds to the reanalyzed score model based on score-based generative models \cite{Song2021}.

Under this parameterization, the reverse process is trained by solving the following optimization problem:
\begin{equation}
\begin{aligned}
\min_{\theta} \mathcal{L}(\theta) :=  &\min_{\theta} \mathbb{E}_{v_0 \sim q(v_0), \epsilon \sim \mathcal{N}(0, I), t} \| \epsilon_t - \epsilon_{\theta}(v_t, t) \|_2^2. \\
% \text{where } v_t  = & \sqrt{\alpha_t} v_0 + (1 - \alpha_t) \epsilon.
\end{aligned}
\end{equation}

The denoising function \( \epsilon_\theta \) estimates the noise \( \epsilon \) added to the noisy input \( v_t \). Trained, we can sample \( v_0 \) from Eq. (\ref{reverse}).

\subsubsection{Forecasting with DDPM}

This work is centered on air pollution forecasting, with implications for broader forecasting problems:
% In this work, we not only focus on air pollution forecasting tasks but also consider the general forecasting problem: 
given a sample \(v_0\) containing missing values, we generate forecasting targets \(v_0^\text{ta} \in \mathcal{V}^\text{ta}\) by leveraging conditional observations \(v_0^\text{co} \in \mathcal{V}^\text{co}\), where both \(\mathcal{V}^\text{ta}\) and \(\mathcal{V}^\text{co}\) are subsets of the sample space \(\mathcal{V}\), and their values vary across samples. The goal of probabilistic forecasting is to approximate the true conditional data distribution \(q(v_0^\text{ta} \mid v_0^\text{co})\) using the model distribution \(p_{\theta}(v_0^\text{ta}\mid v_0^\text{co})\). 
% Typically, all values are used for prediction, i.e., all observed values are set as \(v_0^{co}\) and all forecasting targets are set as \(v_0^{ta}\).
Typically, all values are utilized for prediction, where values considered as observed are set as \(v_0^\text{co}\), and values considered as forecasting targets are set as \(v_0^\text{ta}\).

Next, we explore how to model \(p_{\theta}(v_0^\text{ta} \mid v_0^\text{co})\) using a diffusion model. In the unconditional setting, the reverse process \(p_{\theta}(v_{0:T})\) is used to define the final data model \(p_{\theta}(v_0)\). Therefore, a natural idea is to extend the reverse process in Eq. (\ref{reverse}) to the conditional case:
\begin{equation}
\begin{aligned}
& p_{\theta}(v_{0:T}^{\text{ta}} \mid v_0^{\text{co}}) := p(v_T^{\text{ta}}) \prod_{t=1}^{T} p_{\theta}(v_{t-1}^{\text{ta}} \mid v_t^{\text{ta}}, v_0^{\text{co}}),  v_T^{\text{ta}} \sim \mathcal{N}(0, I),\\
& p_{\theta}(v_{t-1}^{\text{ta}} \mid v_t^{\text{ta}}, v_0^{\text{co}}) := \mathcal{N}\Big(v_{t-1}^{\text{ta}}; \mu_{\theta}(v_t^{\text{ta}}, t \mid v_0^{\text{co}}), \sigma_{\theta}(v_t^{\text{ta}}, t \mid v_0^{\text{co}}) I\Big).
\end{aligned}
\label{eq-con}
\end{equation}

However, existing diffusion models are typically designed for data generation tasks and do not fully utilize the conditional observations \(v_0^\text{co}\). In order to leverage diffusion models for forecasting, previous studies generally adopt two strategies: one approximates the conditional reverse process \(p_{\theta}(v_{t-1}^{\text{ta}} \mid v_t^{\text{ta}}, v_0^{\text{co}})\) with the reverse process in Eq. (\ref{reverse}) \cite{Song2021}; the other uses only historical observations as conditional inputs \cite{Tashiro2021}. The former adds noise to both the target and conditional observations simultaneously, which may compromise the integrity of useful information, while the latter may fail to fully extract historical features, potentially affecting prediction accuracy. Therefore, by directly modeling \(p_{\theta}(v_{t-1}^{\text{ta}} \mid v_t^{\text{ta}}, v_0^{\text{co}})\) without approximations and fully harnessing the wealth of historical features, the predictive accuracy can be markedly enhanced. In the following, we refer to the model defined in Section \ref{DDPM} as the unconditional diffusion model.

\subsection{Spatio-Temporal Physics-Informed Diffusion Models (STeP-Diff)}
\label{3.2}

In this section, we introduce STeP-Diff, a novel spatio-temporal field forecasting method based on diffusion models. This approach not only relies on data-driven forecasts but also strictly adheres to physical laws, ensuring that the forecasted results accurately reflect the actual processes of pollutant diffusion. We provide detailed design insights for two core modules: DeepONet and PDE-Informed Diffusion Models. 
% It is important to note that the application of STeP-Diff is not limited to air pollution forecasting tasks.

% {\color{blue}
% The applicability of STeP-Diff is not restricted to air pollution forecasting. By incorporating a suitable PDE that characterizes the underlying physical dynamics in a given domain, physical constraints can be imposed on the denoising process of the diffusion model. This allows STeP-Diff to be extended to other physically governed spatio-temporal forecasting tasks, such as wildfire spread~\cite{mandel2014recent,mandel2009data} and meteorological evolution~\cite{verma2024climode,fisher2009data}.
% STeP-Diff holds the potential for broader applications. The key to extending STeP-Diff to tasks like wildfire spread~\cite{mandel2014recent,mandel2009data} and meteorological evolution~\cite{verma2024climode,fisher2009data} lies in effectively incorporating task-specific physical constraints into the denoising process, which necessitates further in-depth research.

STeP-Diff has promising potential for broader applications. Extending it to domains such as wildfire spread~\cite{mandel2014recent,mandel2009data} and meteorological evolution~\cite{verma2024climode,fisher2009data} requires the effective integration of task-specific physical constraints into the denoising process. This can be achieved, for instance, by selecting appropriate partial differential equations to guide the denoising process in adhering to physical laws; the implementation of this approach warrants further in-depth investigation. 
% }

\begin{figure*}[ht]
\centering
    \subfigure[DeepONet.]{
        \centering
        \raisebox{0.7cm}{\includegraphics[width=0.32 \columnwidth]{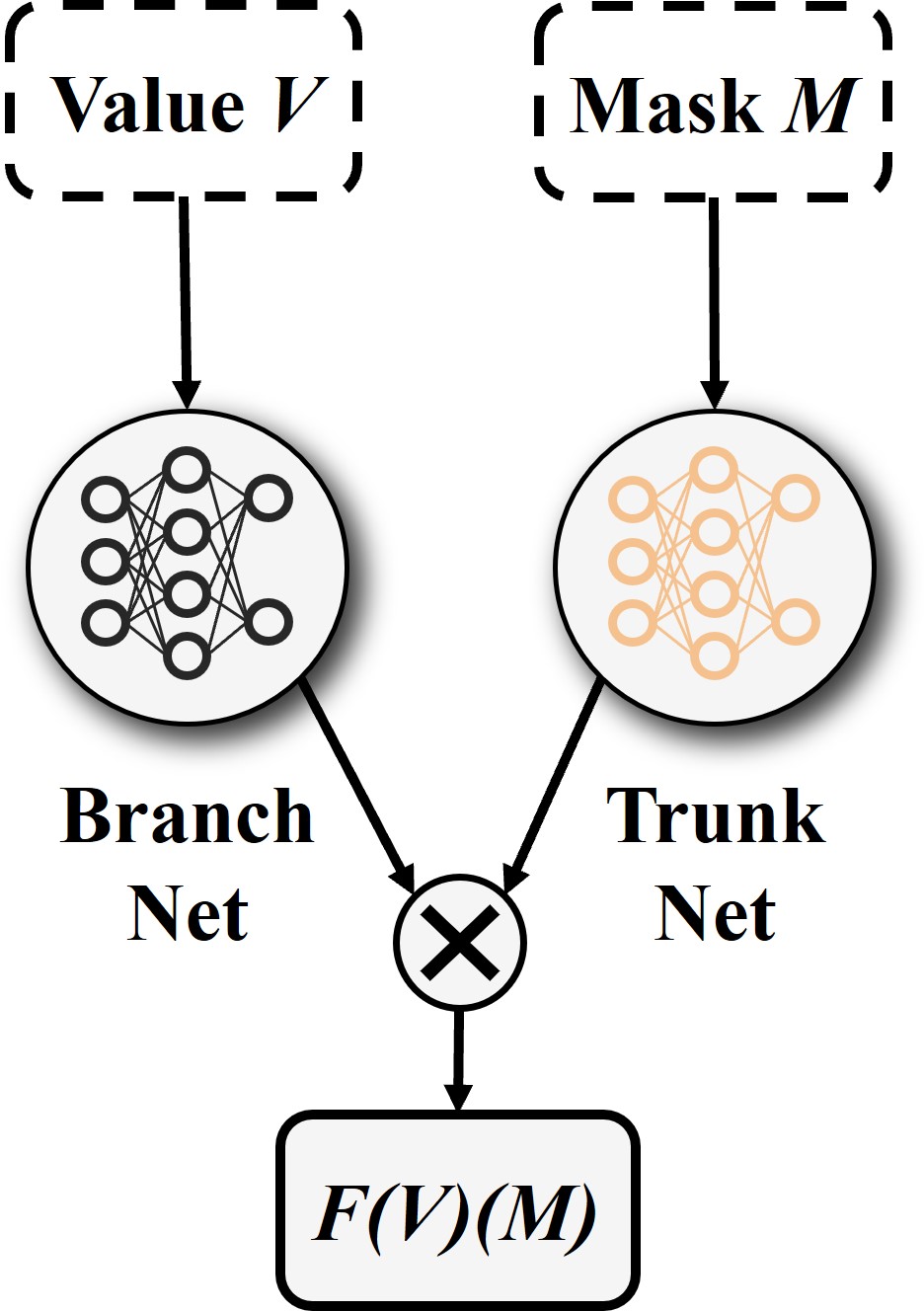}}
    }
    \subfigure[PDE-informed Diffusion Model]{
        \centering
        \includegraphics[width=1.65 \columnwidth]{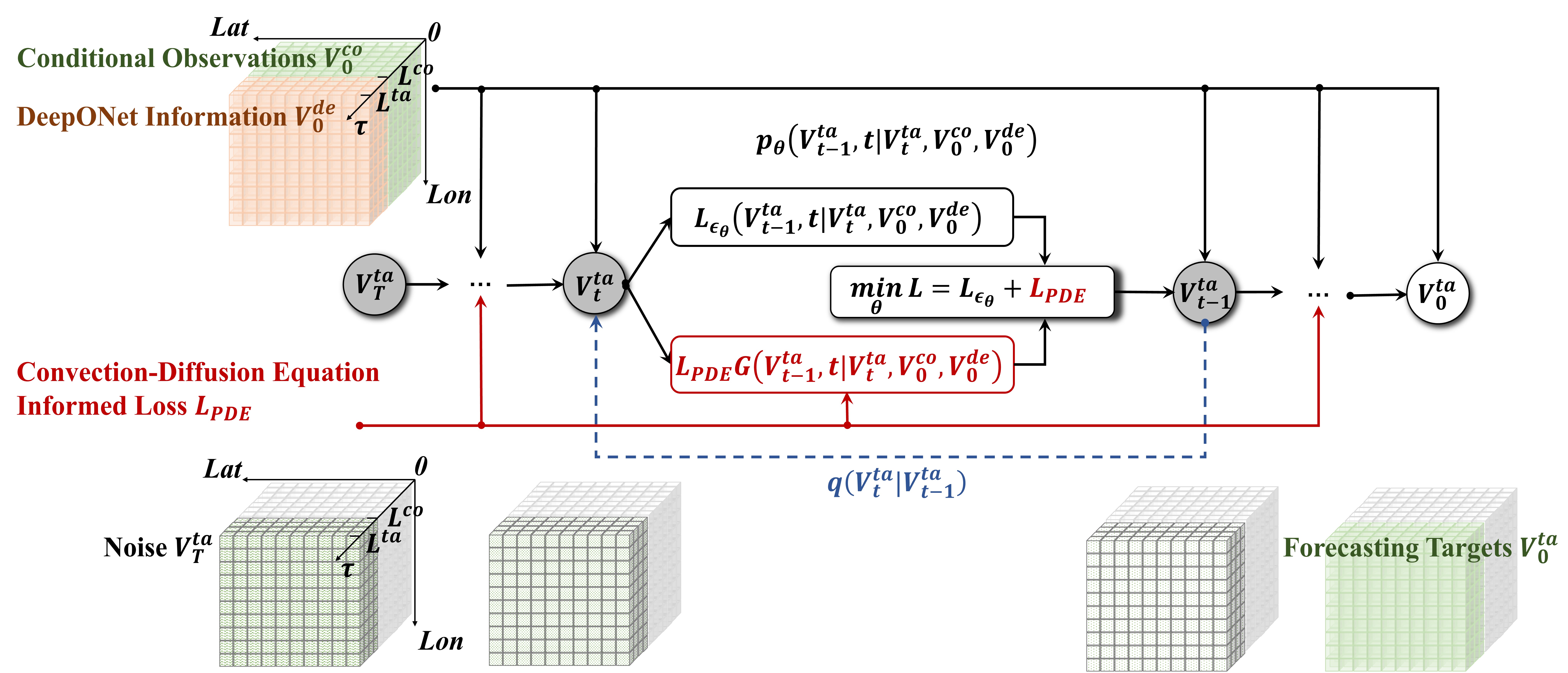}
    }
\caption{Core components of STeP-Diff}
\label{model}
\end{figure*}

\subsubsection{DeepONet}

A neural network with a single hidden layer can accurately approximate any nonlinear continuous operator, revealing the inherent structure and substantial potential of deep neural networks in learning continuous operators or complex systems \cite{Lu2021}.
Inspired by deep operator network (DeepONet), we employ two neural networks, termed the Branch Net and the Trunk Net. The Branch Net encodes the dynamic and incomplete observations \( V \), while the Trunk Net encodes the observation mask \( M \), as illustrated in Fig. \ref{model}(a). Let \( F \) be an operator that takes the input space \( V \) and maps it to the corresponding output space \( F(V) \). For any point \( M \) in the domain of \( F(V) \), the output \( F(V)(M) \) is a real number. Consequently, the network accepts an input composed of two parts, \( V \) and \( M \), and produces the output \( F(V)(M) \): 
\begin{equation}
V^\text{de}=F(V)(M)=\sum \textit{Branch}(V)\cdot \textit{Trunk}(M),
\label{deeponet}
\end{equation}
where both $\textit{Branch}(\cdot)$ and $\textit{Trunk}(\cdot)$ are FNN architecture, $V$ is the observed value, $M \in \{0,1\}$ is the observed mask.

% This module effectively learns the underlying spatio-temporal patterns in historical data, extracting pollutant features at various temporal and spatial points, and thereby provides a robust data foundation for the accurate and rapid forecasting of dynamic and sparse complex spatio-temporal pollutant fields.
This module effectively captures the intrinsic spatial patterns from historical data, distilling pollutant features across multiple spatial locations, thereby establishing a robust foundation for accurate and rapid forecasting of dynamic and incomplete complex spatio-temporal pollutant fields.

\subsubsection{PDE}

The convection-diffusion equation is widely used to describe the propagation of particulate energy and other physical quantities, including the dispersion of air pollutants. The equation is given by:
\begin{equation}
\frac{\partial V}{\partial t} = \nabla \cdot (K \nabla V) - \nabla V \cdot (p V) + R,
\label{cde}
\end{equation}
where the variable of interest $V = V(x,y,z,\tau)[\text{kg/m}^3]$ represents the air pollutant concentration at position $(x,y,z)$ and time $\tau$. The wind velocity is denoted as $p=(p_x,p_y)$[m/s] and the diffusion coefficient is represented by $K=\text{diag} (K_x,K_y,K_z)[\text{m}^2\text{/s}]$. Additionally, $R [\text{kg/m}^3\text{s}]$ accounts for environmental factors such as air pollution sources.

In Eq. (\ref{cde}), the time evolution of air pollution $\frac{\partial V}{\partial \tau}$ is influenced by three factors: 
(1) pollutant flux due to diffusion, $\nabla \cdot (K \nabla V)$; 
(2) pollutant flux due to wind, $ - \nabla V \cdot (p V)$;
(3) local creation or destruction of air pollution, $R$. 
This equation is well-suited for estimating fine particulate air pollution, as it reveals the dynamics of air evolution and describes the relationships between quantities at different times and locations. When the coverage of observations is incomplete, these relationships assist in inferring air pollution values in unobserved regions.

However, due to the computational expense of this model, appropriate assumptions need to be made to reduce complexity while maintaining accuracy. Additionally, because of its partial differential form, the current expression $V^{\tau}$ does not directly allow the derivation of air pollution at the next time step, $V^{\tau +1}$.
% Therefore,  the previous work in our laboratory \cite{Chen2020} rederived the partial differential equation (PDE) in Eq. (\ref{cde}) to suit physics-guided state evolution estimation:
% Our laboratory's previous work rederived PDE in for physics-informed state evolution estimation. 
Our laboratory's previous work rederived the PDE for physics-informed state evolution estimation.
For detailed derivations, please refer to \cite{Chen2020}. 

Let $G$ be an operator, the derived result is as follows:
\begin{equation}
\begin{aligned}
% & V^{\tau + {\color{blue}\Delta \tau}} = G(V^{\tau}) = B V^{\tau} + C S^{\tau}, \\
& V^{\tau + \Delta \tau} = G(V^{\tau}) = B V^{\tau} + C S^{\tau}, \\
\text{where }&B := \exp(\Delta \tau A),
C  := \left( \exp(\Delta \tau A) - I \right) A^{-1}, \\
& A((x,y),(x,y+1)) = \frac{K}{n^2} - \frac{P_y(x,y)}{n},\\
& A((x,y),(x+1,y)) = \frac{K}{n^2} - \frac{P_x(x,y)}{n},\\
& A((x,y),(x,y-1)) = A((x,y),(x-1,y)) = \frac{K}{n^2},\\
& A((x,y),(x,y)) = -\frac{4K}{n^2} - \frac{1}{n} [ P_x(x+1,y) - \\ 
& \hspace{1.5cm} 2P_x(x,y) + P_y(x,y+1) - 2P_y(x,y)], 
\end{aligned}
\label{PDE}
\end{equation}
% {\color{blue} and $V^\tau$ is the pollution measurement at time $\tau$};
% $S^\tau$ is derived from the air pollution data collected before time slice $\tau$ using least square \cite{chen2018pga}; $K$ is the diffusion coefficient; and $P$ is the wind speed.
with $V^\tau$ representing the pollution measurement at time $\tau$;
$S^\tau$ is obtained from air pollution prior to $\tau$ using least squares \cite{chen2018pga}; $K$ is the diffusion coefficient; $P$ is the wind speed.

% {\color{blue}
Based on this, we adopt a time-slicing discretization scheme to represent the temporal sequence, where each index $l$ corresponds to a discrete time slice, and the interval between $l$ and $l+1$ is a fixed time step $\Delta\tau$. This leads to the following discrete form of the air dispersion model:
\begin{equation}
 V^{l+1} = G(V^{l}) = B V^{l} + C S^{l},
\label{PDE-l}
\end{equation}
% }
where $V^{l}$ is the pollution value at discrete time slice $l$.

\begin{algorithm*}[t]
\SetAlgoLined
\SetKwInOut{Input}{Input}
\SetKwInOut{Output}{Output}
\Input{Observed data $V_0=\{v_0^{1:L,1:X,1:Y}\} \in \mathcal{R}^{L\times X\times Y}$, 
% where $L$ is the length of the time silces, $X$ is the number of longitude coordinate points, and $Y$ is the number of latitude coordinate points; 
observed mask $M_0=\{m_0^{1:L,1:X,1:Y}\} \in \{0,1\}^{L\times X\times Y}$, 
% where $m_0^{l,x,y}=0$ if $v_0^{l,x,y}$ is missing and $m_0^{l,x,y}=1$ if $v_0^{l,x,y}$ is observed; 
the number of iteration $N_{\text{iter}}$; 
the sequence of noise levels $\{\alpha_t\}$, DeepONet operator \{$F(\cdot)(\cdot)$\} accoroding to Eq. (\ref{deeponet}), the computationally practical form PDE operator \{$G(\cdot)$\} accoroding to Eq. (\ref{PDE-l}).}
\Output{Trained denoising function $\epsilon_\theta$}
\BlankLine
\For{$i = 1$ \textbf{to} $N_{\text{iter}}$}{
    Sample $t \sim \text{Uniform}(\{1, \dots, T\})$ \;
    Split time series $L$ into $L_1$ and $L_2$ such that $L = L_1 + L_2$, and set the last $L_2$ steps as forecasting targets \;
    Partition $\{V_0, M_0\}$ into two sets: conditional historical observations $\{V_0^\text{co}, M_0^\text{co}\} \in \{ \mathcal{R}^{L_1\times X\times Y}, \{0,1\}^{L_1\times X\times Y}\}$ and forecasting targets $\{V_0^\text{ta}, M_0^\text{ta}\} \in \{ \mathcal{R}^{L_2\times X\times Y}, \{0,1\}^{L_2\times X\times Y} \}$ \;
    Forecast $\{V_0^\text{de}\} = F(V_0^\text{co})(M_0)$ where $V_0^\text{de} = \{{v_0^\text{de}}^{1:L_2,1:X,1:Y}\}\in \mathcal{R}^{L_2\times X\times Y}$ for $L_2$ steps using DeepONet \;
    Sample noise $\epsilon_t \sim \mathcal{N}(0, I) \in \mathcal{R}^{L_2 \times X \times Y}$ \;
    Compute noisy targets: $V_t^\text{ta} = \{{v_0^\text{ta}}^{1:L_2,1:X,1:Y}\} = \sqrt{\alpha_t} V_0^\text{ta} + (1 - \alpha_t)\epsilon, \epsilon=\sum_{j=1}^t {\epsilon_j} $ \;
    Update parameters with gradient step on $\hat{\epsilon}_\theta = \epsilon_\theta(v_t^{\text{ta}}, t \mid v_0^{\text{co}}, m_0^\text{co}, v_0^\text{de})$ according to Eq. (\ref{loss_n}):
    \[
    \nabla_\theta \left\{ \left\| \epsilon_t^{\{1:L_2,1:X,1:Y\}} - \hat{\epsilon}_\theta^{\{1:L_2,1:X,1:Y\}} \right\|^2 + \omega \cdot \left\| \epsilon_t^{\{2:L_2,1:X,1:Y\}} - G(\hat{\epsilon}_\theta^{\{1:L_2-1,1:X,1:Y\}} )\right\|^2 \right\}
    \] 
}

\caption{Training of STeP-Diff.}
\label{train}
\end{algorithm*}
\begin{algorithm*}[t]
 \SetAlgoLined
 \SetKwInOut{Input}{Input}
 \SetKwInOut{Output}{Output}
 \Input{Sample observations $V_0^\text{co} \in \mathcal{R}^{L_1\times X\times Y}$; sample mask $M_0^\text{co} \in \{0,1\}^{L_1\times X\times Y}$; forecasting mask $M_0^\text{ta} \in \{0\}^{L_2\times X\times Y}$; trained denoising function $\epsilon_\theta$; DeepONet operator \{$F(\cdot)(\cdot)$\} according to Eq. (\ref{deeponet}).}
 \Output{Forecasting values $V_0^{\text{ta}} \in \mathcal{R}^{L_2\times X\times Y}$}
\BlankLine
Set $L_1+L_2$ timesteps as one time series $L$, and denote the last $L_2$ steps' values as forecasting targets \
$\{V_0,M_0\} \in \{ \mathcal{R}^{L\times X\times Y}, \{0,1\}^{L\times X\times Y}\}$ is composed of two sets: observations$\{V_0^\text{co},M_0^\text{co}\} \in \{ \mathcal{R}^{L_1\times X\times Y}, \{0,1\}^{L_1\times X\times Y}\}$ 
and targets $\{V_0^\text{ta},M_0^\text{ta}\} \in \{ \mathcal{R}^{L_2\times X\times Y}, \{0\}^{L_2\times X\times Y}\}$ \;
$V_0^\text{de} = F(V_0^\text{co})(M_0)$ where $V_0^\text{de} \in \mathcal{R}^{L_2 \times X \times Y}$ represents the data forecasted by DeepONet for $L_2$ steps \;
$V_t^{\text{ta}} \sim \mathcal{N}(0, I)$, where the dimension of $V_t^{\text{ta}}$ corresponds to the forecasting indices of $V_0^\text{ta}$\;

 \For{$t = T$ \textbf{to} 1}{
     Sample $V_{t-1}^\text{ta}$ using Eq. (\ref{con_n}) and (\ref{para_n}) \;
 }
 \caption{Forecasting (Sampling) with STeP-Diff.}
 \label{sample}
\end{algorithm*}

\subsubsection{PDE-Informed Diffusion Models}
Based on Eq. (\ref{deeponet}), the measurement values $V^\text{co}$ from the historical observation period $L^\text{co}$ are input into DeepONet, which outputs the preliminary predicted values $V^\text{de}$ for the target prediction period $L^\text{ta}$. To more effectively incorporate spatio-temporal features, we propose an innovative approach by using both the historical observations $V^\text{co}$ and the output from DeepONet $V^\text{de}$ as joint conditional inputs, as shown in Fig. \ref{model}(b). This method not only enhances the model's understanding of spatio-temporal features but also leverages the denoising advantage of diffusion models to further improve the learning capability for data distributions. Therefore, the conditional distribution in Eq. (\ref{eq-con}) can be rewritten as:
\begin{equation}
\begin{aligned}
& p_{\theta}(V_{0:T}^{\text{ta}} \mid V_0^\text{info})  :=  p(V_T^{\text{ta}}) \prod_{t=1}^{T} p_{\theta}(V_{t-1}^{\text{ta}} \mid V_t^{\text{ta}}, V_0^\text{info}),  \\
& p_{\theta}(V_{t-1}^{\text{ta}} \mid V_t^{\text{ta}}, V_0^\text{info})  :=  \mathcal{N}(V_{t-1}^{\text{ta}}; \mu_{\theta}(V_t^{\text{ta}}, t \mid V_0^\text{info}), \\
& \hspace{5cm} 
\sigma_{\theta}(V_t^{\text{ta}}, t \mid V_0^\text{info}) I),
\end{aligned}
\label{con_n}
\end{equation}
where $V_0^\text{info}=\{V_0^{\text{co}},M_0^{\text{co}},V_0^{\text{de}}\}$ represents the complete set of conditional information provided to the Diffusion model, 
including the observed measurements $V_0^{co}$, the observed measurement mask $M^{co}$, and the prediction output of DeepONet $V_0^{de}$.
The forecasting targets are initialized as Gaussian noise, denoted as $ V_T^{\text{ta}} \sim \mathcal{N}(0, I).$

To model the conditional distribution  $p_\theta (V_{t-1}^{\text{ta}} \mid V_t^{\text{ta}}, V_0^\text{info})$ without approximations, we focus on the conditional diffusion model of the reverse process in Eq. (\ref{con_n}). Specifically, we extend the parameterization of the DDPM in Eq. (\ref{reverse}) to the conditional setting and define the conditional denoising function \( \epsilon_{\theta}:(V^\text{ta} \times \mathbb{R} \mid V^\text{info}) \rightarrow V^\text{ta} \), where $ V^\text{info}= \{ V^\text{co}, M^\text{co}, V^\text{de} \}$ serves as conditional inputs. Additionally, we parameterize \( \epsilon_{\theta} \) as follows:
\begin{equation}
\begin{aligned}
&\mu_{\theta}(V_t^{\text{ta}}, t \mid V_0^\text{info}) = \mu^{\text{DDPM}}(V_t^{\text{ta}}, t, \epsilon_{\theta}(V_t^{\text{ta}}, t \mid V_0^\text{info})),\\
&\sigma_{\theta}(V_t^{\text{ta}}, t \mid V_0^\text{info}) = \sigma^{\text{DDPM}}(V_t^{\text{ta}}, t),
\end{aligned}
\label{para_n}
\end{equation}
where \( \mu^\text{DDPM} \) and \( \sigma^\text{DDPM} \) are the functions defined in Section \ref{DDPM}. Based on \( \epsilon_{\theta} \) and data \( V_0 \), we can sample the forecasting targets \( V_0^\text{ta} \) using the reverse process described in Eq. (\ref{con_n}) and Eq. (\ref{para_n}). In the sampling process, we set all observed \( V_0 \) as the conditional observations \( V_0^\text{co} \) and the pollution values to be predicted as the prediction targets \( V_0^\text{ta} \). It is important to note that, without the conditional input, the conditional model degenerates into an unconditional model, which can be used for data generation tasks.

However, due to the dependence of the DDPM sampling algorithm on generating data samples from \( V^\text{ta}_T \sim \mathcal{N}(0, I) \) via a Markov chain, the randomness of \( V_T \) makes the data generation process difficult to control precisely. Therefore, under conditions of incomplete and noisy measurements, achieving efficient and accurate predictions is challenging. 
To address this issue, we impose physical constraints on the reverse diffusion process of noise using PDE, guiding it to gradually adhere to the inherent diffusion dynamics embedded within the equations.

% Specifically, during the training of \( \epsilon_{\theta} \), the network's estimated noise must not only approximate the randomly added noise but also ensure that the noise evolves according to the PDE's physical diffusion laws between adjacent time silces. 
% {\color{blue}
% Specifically, we introduce a PDE-based regularization term to enforce that the evolution of the noise sequence $\epsilon$ also follows the physical laws governing pollutant dispersion. This constraint is formulated as $\epsilon^{\{l+1,x,y\}} = G(\epsilon^{\{l,x,y\}})$, and can be further expressed in a sequential form as $\epsilon^{\{2:L,1:X,1:Y\}} = G(\epsilon^{\{1:L-1,1:X,1:Y\}})$.
% }

Before delving into the PDE-based physical constraints, we first present a theorem derived from the noise addition process of the diffusion model:
\begin{theorem}
Define the noise term as:
\begin{equation}
\epsilon^{l+1}_t=G(\epsilon^l_t)=B\epsilon_t^l+C_tS^l,
\label{noise_PDE}
\end{equation}
where $G(\cdot)$ is the PDE evolution operator defined in Eq. (\ref{PDE-l}). Then, the pollutant concentration satisfies:
\begin{equation}
{V_0}^{l+1}=G({V_0}^l).
\label{pollution_PDE}
\end{equation}
\end{theorem}
\begin{proof}
See Appendix~\ref{proof} for proof.
\end{proof}
Based on Eq. (\ref{noise_PDE}), we incorporate the $L_\text{PDE}$ regularization term into the loss function to enforce PDE constraints at each denoising step, thereby ensuring that the generated pollution concentration sequences also adhere to the convection-diffusion equation at the physical level (i.e., Eq. (\ref{pollution_PDE})).

Unlike the training of the unconditional model in Section \ref{DDPM}, for a given set of conditional observations \( V_0^\text{co} =\{{v_0^\text{co}}^{1:L_1,1:X,1:Y}\}\), observed mask  \( M_0^\text{co} = \{{m_0^\text{co}}^{1:L_1,1:X,1:Y}\} \), DeepONet output \( V_0^\text{de} =\{{v_0^\text{de}}^{1:L_2,1:X,1:Y}\} \),  and forecasting target \( V_0^\text{ta} =\{{v_0^\text{ta}}^{1:L_2,1:X,1:Y}\} \), we sample the noise target as follows: \(
V_t^\text{ta}  =\{{v_t^\text{ta}}^{1:L_2,1:X,1:Y}\} = \sqrt{\alpha_t} V_0^\text{ta} + (1 - \alpha_t) \epsilon,\)
and train \( \epsilon_{\theta} \) by minimizing the following loss function:
\begin{equation}
\begin{aligned}
    & \min_{\theta} L(\theta)  := \min_{\theta}E_{v_0, \epsilon_t \sim \mathcal{N}(0, I)} \Big(  L_{\epsilon_\theta} +  \omega \cdot L_{PDE} \Big),\\
    &L_{\epsilon_\theta} =\|  \epsilon_t^{\{1:L_2,1:X,1:Y\}} -\epsilon_\theta(v_t^{\text{ta}}, t \mid v_0^{\text{info}})^{\{1:L_2,1:X,1:Y\}} \|^2,\\
    & L_\text{PDE} = \| \epsilon_t^{\{2:L_2,1:X,1:Y\}} - G(\epsilon_\theta(v_t^{\text{ta}}, t \mid v_0^{\text{info}})^{\{1:L_2-1,1:X,1:Y\}}) \|^2,
\end{aligned}
\label{loss_n}
\end{equation}
where $v_0^\text{info}=\{v_0^{\text{co}}, m_0^\text{co}, v_0^\text{de}\}$, $G(\cdot)$ represents the PDE operator defined in Eq. (\ref{PDE-l}), the noise term \( \epsilon \) shares the same dimensions as the forecasting target \( v_0^\text{ta} \), and $\omega$ denotes the weight parameter.

% By introducing PDE-based physical constraints, this model not only facilitates data-driven prediction but also rigorously adheres to physical laws, ensuring that the predicted results reflect the actual pollution diffusion process. 
% {\color{blue}
% By introducing the $L_\text{PDE}$ regularization term, we effectively impose partial differential equation constraints on each denoising step, ensuring that the generated sequences adhere to the convection-diffusion equation at the physical level. This guarantees that the predictions maintain physical consistency and interpretability.
% By incorporating the $L_\text{PDE}$ regularization term, we effectively enforce partial differential equation constraints at each denoising step, ensuring that the generated sequences rigorously adhere to the convection-diffusion equation at the physical level. This enhances the physical consistency and interpretability of the predictions. A detailed derivation is provided in Appendix~\ref{proof}.
% }

However, this training procedure still faces a challenge: since precise future pollution values (i.e., ground-truth prediction values) are not available in real-world applications, we cannot directly determine the division of observed measurement \( V_0^\text{co} \) and forecasting targets \( V_0^\text{ta} \) from the training sample \( V_0 \).
To address this, we adopt a self-supervised learning strategy, inspired by the concept of masked language modeling. Specifically, for a given sample \( V_0 \), we partition the observations into two parts, designating one part as the conditional observation \( V_0^\text{co} \) and the other as the prediction target \( V_0^\text{ta} \). We then sample the noise target \( V_T^\text{ta} \) and train \( \epsilon_{\theta} \) by solving Eq. (\ref{loss_n}). The specific training and sampling algorithms will be detailed in the following Section \ref{alg}.

\subsection{Implementation of STeP-Diff for Forecasting}
\label{3.3}

\subsubsection{The Architecture of $\epsilon_{\theta}$}

% We adopt the network architecture proposed in CSDI \cite{Tashiro2021}, which is designed to capture temporal and feature dependencies in multivariate time series. 
% We adopt the network architecture proposed in CSDI \cite{Tashiro2021}, which is specifically designed to capture both temporal dependencies and interactions among features in multivariate time series, ensuring a comprehensive understanding of the underlying dynamics.
% Unlike convolution-based architectures, this model incorporates a bi-dimensional attention mechanism in each residual layer to enhance the modeling of complex patterns. Specifically, the network employs a Temporal Transformer layer, which takes the tensor of each feature as input to learn temporal dependencies, and a Feature Transformer layer, which processes the tensor at each time step to model dependencies across different features.  
We adopt the network architecture proposed in CSDI \cite{Tashiro2021}, which is specifically designed to capture both temporal dependencies and interactions among features in multivariate time series, ensuring a comprehensive understanding of the underlying dynamics.
Unlike convolution-based architectures, this model incorporates a bi-dimensional attention mechanism in each residual layer to enhance the modeling of complex patterns. The network employs a Temporal Transformer layer, which takes the tensor of each feature as input to learn temporal dependencies, and a Feature Transformer layer, which processes the tensor at each time step to model dependencies across different features.

Notably, although the length of each time series \( L \) may vary, as mentioned in Section \ref{math}, the self-adaptive nature of the attention mechanism allows the model to handle sequences of different lengths. During batch training, zero-padding is applied to each sequence to align their lengths, facilitating efficient parallel computation.

\subsubsection{Training \& Sampling Algorithm}
\label{alg}

In Algorithm \ref{train}, we provide a detailed description of the training procedure for STeP-Diff, while Algorithm \ref{sample} outlines the process for forecasting (sampling) using the trained STeP-Diff model.

During the training phase, the model takes as inputs observed data $V_0$, observed mask $M_0$, the number of iterations $N_\text{iter}$, the sequence of noise levels $\{\alpha_t\}$, DeepONet operator $F$, and PDE operator $G$. 
Initially, the observed data and mask $\{V_0, M_0\}$ are partitioned into conditional observations $\{V_0^\text{co}, M_0^\text{co}\}$ and forecasting targets$\{V_0^\text{ta}, M_0^\text{ta}\}$.
The conditional observations $V_0^\text{co}$, along with the complete observed mask $M_0$, are then fed into the DeepONet operator $F$ to obtain preliminary predictions $V_0^\text{de}$. Subsequently, noisy targets $V_t^\text{ta}$ are obtained by progressively adding noise to the prediction targets $V_0^\text{ta}$. Finally, the conditional observations $V_0^\text{co}$, conditional masks $M_0^\text{co}$, DeepONet outputs $V_0^\text{de}$, and noisy targets $V_t^\text{ta}$ are jointly input the denoising network $\epsilon_\theta$ for training, optimized by minimizing a loss function that incorporates PDE-based physical constraints based on $G$, thus learning the final denoising function.

During the sampling phase, the model receives as inputs sample observations $V_0^\text{co}$, sample mask $M_0^\text{co}$, forecast mask $M_0^\text{ta}$, trained denoising network $\epsilon_\theta$, and DeepONet operator $F$. 
First, the complete data mask $M_0$ is constructed by merging the sample mask $M_0^\text{co}$ with the forecast mask $M_0^\text{ta}$. Next, the sample observations $V_0^\text{co}$, combined with the complete mask $M_0$, are input into the DeepONet operator $F$ to generate preliminary predictions $V_0^\text{de}$. Subsequently, noisy targets $V_t^\text{ta}$ are sampled from a normal distribution that conforms to the dimensionality of the prediction targets $V_0^\text{ta}$. Finally, the sample observations $V_0^\text{co}$, sample mask $M_0^\text{co}$, DeepONet outputs $V_0^\text{de}$ and noisy targets $V_t^\text{ta}$ are provided to the denoising network $\epsilon_\theta$, where a sampling procedure is executed to produce the final prediction results $V_0^\text{ta}$.

\section{Evaluation}

\label{eval}

In this section, we evaluate the ability of our system to address the challenges of incomplete and time-varying data coverage, aiming to achieve accurate fine-grained air pollution predictions. We first introduce in Section \ref{Dep} how our system is deployed to collect urban-scale data for evaluation. 
Meanwhile, a statistical analysis of the data coverage is presented in Appendix \ref{A2}.
% Then, we conduct a statistical analysis of the data coverage in Section \ref{Sta}. 
Finally, in Section \ref{Sys},  we conduct extensive experiments to demonstrate the superiority of our algorithm. 

\subsection{Deployment \& Evaluation Setup}
\label{Dep}

To collect air pollution data for performance evaluation, portable sensing devices were designed for city-scale deployment, with details provided in Section \ref{por}. The portable sensing devices were deployed in two cities for 14 days, as described in Section \ref{cit}. Additionally, Section \ref{eva} provides the details of the evaluation setup.

\subsubsection{Portable Sensing Devices}
\label{por}
The portable sensing device is designed to collect air pollution data and transmit it to the cloud for air pollution estimation. The design principle of the equipment is shown in Fig. \ref{fig:mobile}(a) \cite{Chen2020, liu2024mobiair}. The sensing device consists of four functional modules: sensing module, control module, computation module, and power module.

\paragraph{Sensing Module} This module is responsible for collecting air pollution data, including GPS sensors that record the location and time, and air pollutant sensors that measure the concentrations of six pollutants (PM$_{2.5}$, PM$_{10}$, O$_3$, SO$_2$, NO$_2$, CO). The sensor units used in this module and their corresponding accuracy are listed in Table \ref{sensor}. 
% Data is collected every three seconds.

\paragraph{Control Module} This module primarily comprises a microcontroller that manages the sensing, communication, and processing functions within the sensing hardware.

\paragraph{Computation Module} This module transmits the sensed data to the cloud over 3G/4G and WiFi networks, facilitating real-time online inference.

\paragraph{Power Module} 
% This module features a power management integrated circuit that stabilizes the vehicle's power supply, ensuring a reliable and consistent output for the sensing module.
This module features a power management integrated circuit that stabilizes the vehicle's power supply, ensuring consistent output for the sensing module.

\subsubsection{City-Scale Deployment}
\label{cit}

\begin{figure*}[ht]
    \centering
    \subfigure[Principle of portable sensing devices.]{
        \centering
        \includegraphics[height=3.3 cm]{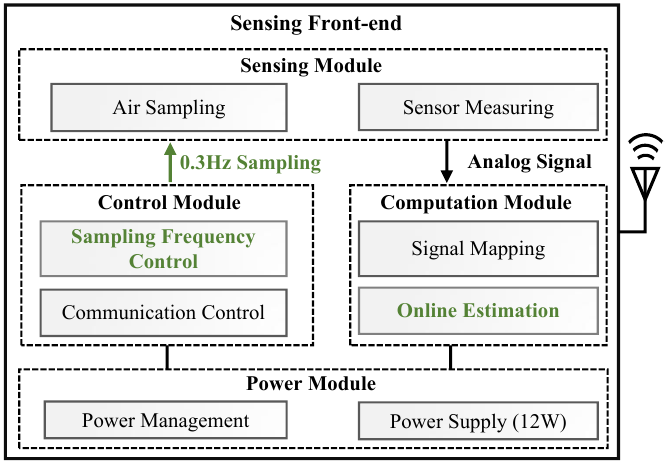}
    }
    \subfigure[Installation on cars in Nanjing.]{
        \centering
        \includegraphics[height=3.3 cm]{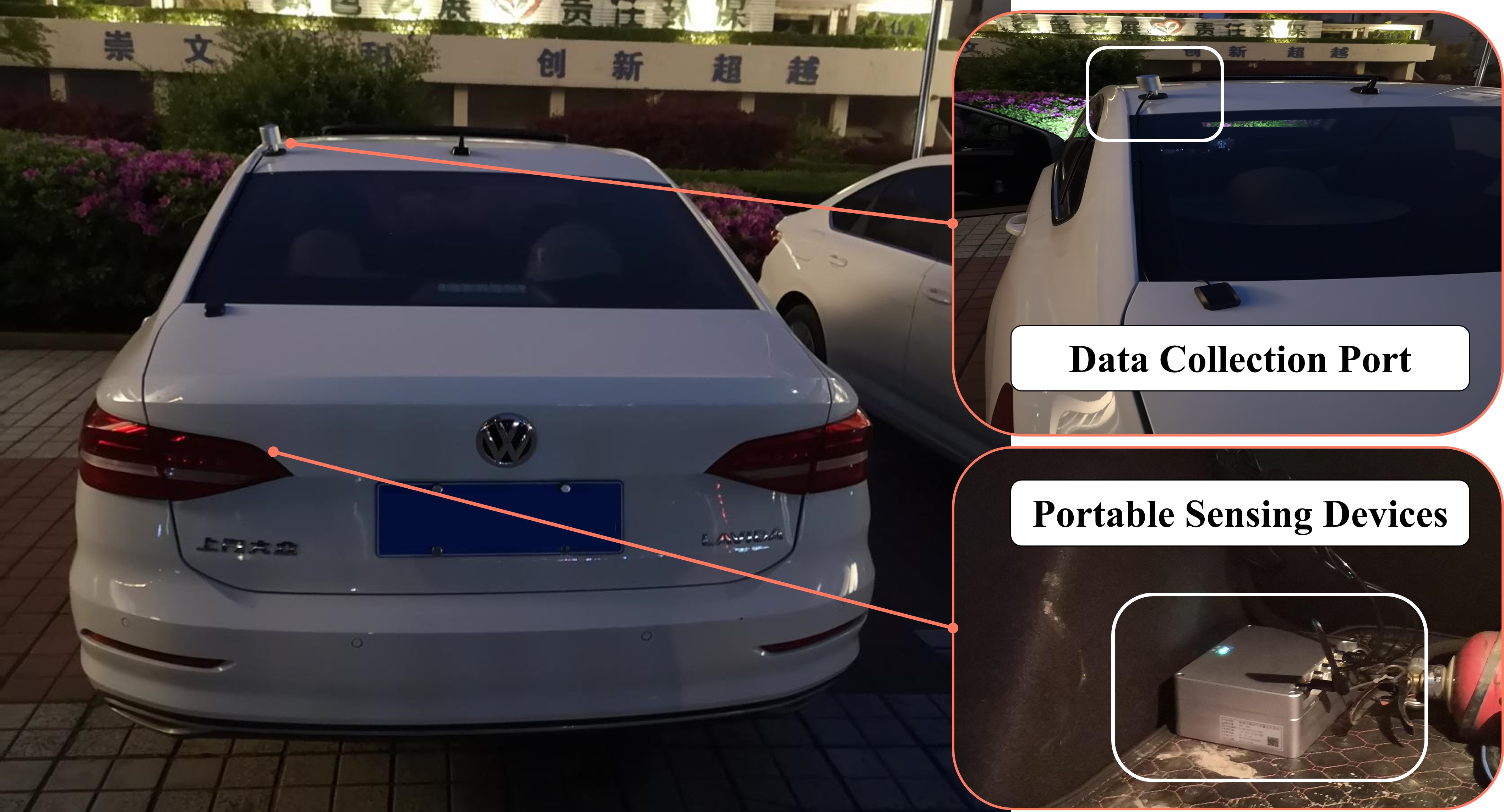}
    }
    \subfigure[Installation on buses in Changshu.]{
        \centering
        \includegraphics[height=3.3 cm]{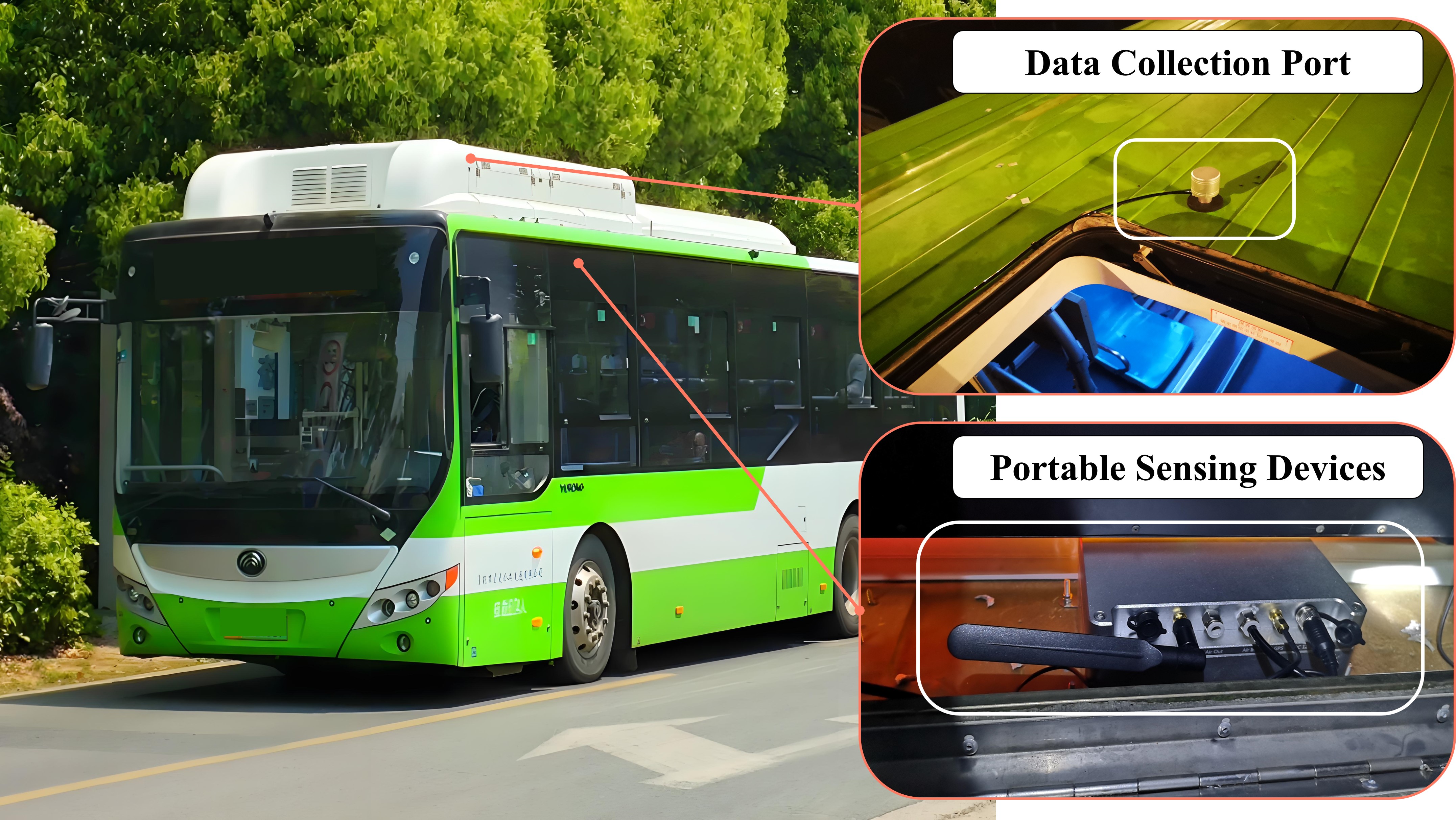}
    }
    \caption{Principle of the self-designed portable sensing devices and their installation on mobile platforms.}
    \label{fig:mobile}
\end{figure*}

To collect urban-scale air pollution data for performance evaluation, we deployed portable sensing devices on 27 cars in Nanjing and 32 buses in Changshu, both located in Jiangsu Province, China \cite{li2024quest}. These devices operated for 14 days to gather experimental data. The installation setup is shown in the Fig. \ref{fig:mobile}(b)(c). The vehicles operated without any interference, ensuring data collection under realistic conditions.

\begin{figure}[t!]
  \centering
  \subfigure[Nanjing.]{
    \centering
    \includegraphics[height=0.36\linewidth]{samples/Fig/NJ_Trace.jpg}
  }
  \hfill
  \subfigure[Changshu.]{
    \centering
    \includegraphics[height=0.4\linewidth]{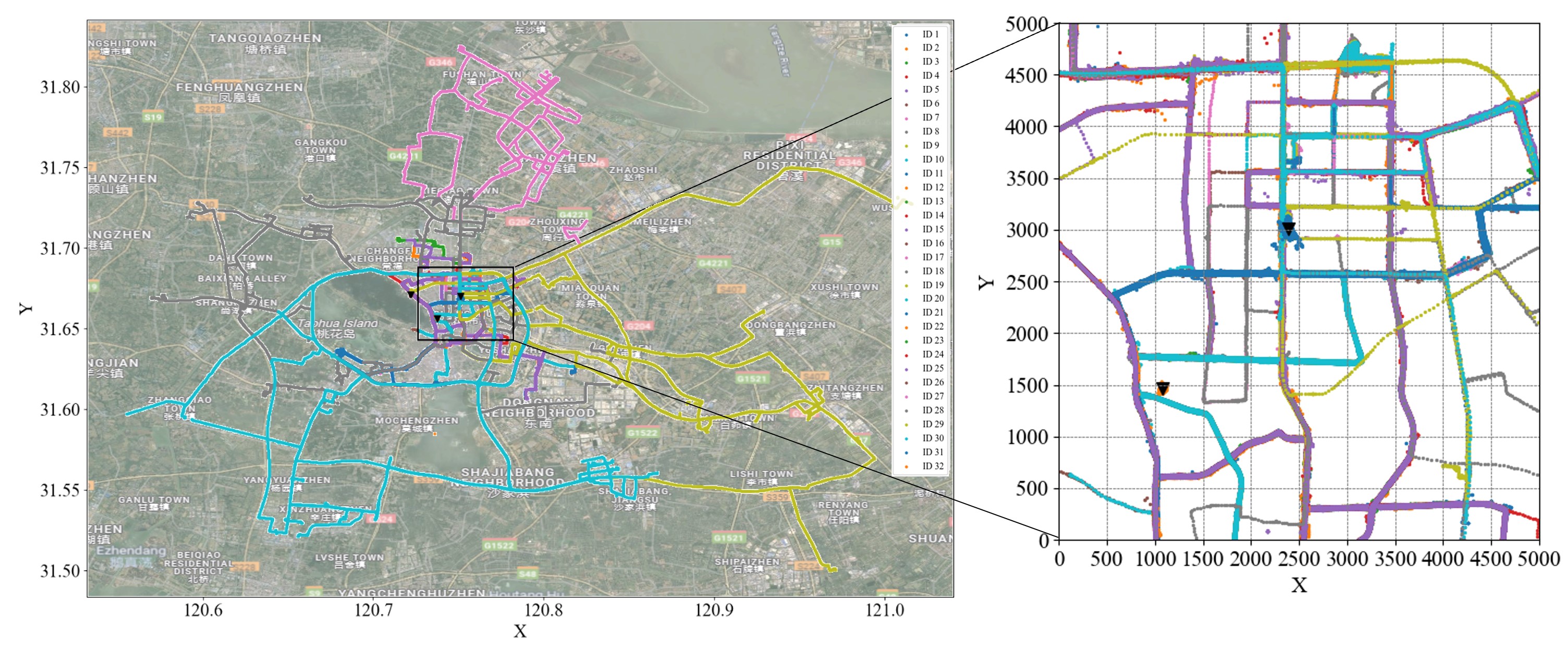}
  }
  \caption{The operational trajectory of the mobile devices and the locations of fixed monitoring stations. The lines in different colors represent the trajectories of different vehicles, while the black triangles indicate the locations of the fixed monitoring stations.}
  \label{fig:trace}
\end{figure}

The primary geographic coverage of the mobile devices in Nanjing spans the latitude and longitude range of [118.600:118.950, 31.900:32.075], with the time range from April 17, 2023, 00:00 to April 30, 2023, 23:59. In Changshu, the coverage area is [120.500:121.000, 31.500:31.850], with the time span from June 7, 2021, 00:00 to June 22, 2021, 23:59, as illustrated in the left part of Fig. \ref{fig:trace}(a)(b).
% In addition to mobile sensing measurements, Nanjing had 23 official monitoring stations, while Changshu had 3,  as indicated by the black triangles in the figure. These official stations provided reference data, allowing further validation of the reliability of the mobile platform’s assessment results.

To further analyze the spatio-temporal evolution patterns of air pollution, we selected the 5km$\times$5km sub-region with dense mobile device trajectories in both cities for in-depth study, covering a 14-day period for each.
The geographical coordinates of the selected sub-region in Nanjing range from [118.760:118.813, 32.000:32.045], and the sub-region in Changshu ranges from [120.726:120.782, 31.643:31.688]. These areas are zoomed in on the right side of Fig. \ref{fig:trace}(a)(b). For ease of interpretation, we converted the geographical coordinates into kilometers and set the bottom-left corner of each sub-region as the origin. In these sub-regions, Nanjing has 10 monitoring stations and Changshu has 2, providing official data support for our evaluation.

\subsubsection{Evaluation Setup}
\label{eva}

This subsection provides an overview of the evaluation setup, including data description, ground truth, system parameters, performance metrics, and baselines.

\paragraph{Data Description}
% \textbf{Data Description}
% The data collected and transmitted to the cloud includes the following format: device ID, timestamp, longitude, latitude, PM$_{2.5}$ concentration, PM$_{10}$ concentration, O$_3$ concentration, SO$_2$ concentration, NO$_2$ concentration, and CO concentration. All pollutant concentrations are expressed in $\mu g/m^3$. Since PM$_{2.5}$ has been shown to be easily absorbed by the human respiratory system, potentially leading to respiratory and even blood-related diseases, we will focus on evaluating PM$_{2.5}$ in the subsequent experiments.
We collected device ID, timestamp, longitude, latitude, PM$_{2.5}$ concentration, PM$_{10}$ concentration, O$_3$ concentration, SO$_2$ concentration, NO$_2$ concentration, and CO concentration, and transmitted this data to the cloud. All pollutant concentrations are measured in $\mu g/m^3$. Since PM$_{2.5}$ has been shown to be easily inhaled by the human respiratory system, potentially leading to respiratory and even blood-related diseases, we will focus on evaluating PM$_{2.5}$ in the subsequent experiments.

\paragraph{Ground Truth}
The limited number of monitoring stations provides ground truth data for only a small portion of the area, which may lead to biased evaluation results. To address this issue, we evaluate using measurements from both monitoring stations and mobile devices. Given that mobile device measurements are more abundant than those from monitoring stations, we train the model using only the mobile device measurements during the training phase. In the evaluation phase, we compute performance metrics by using both the monitoring station and mobile device measurements as ground truth.
This approach not only enables us to evaluate prediction accuracy over a larger area but also validates the reliability of the mobile device measurements. Although unobserved sub-regions exist in each time slice, the high mobility and random selection strategy of sensing platform ensured that most accessible sub-regions were validated during the 14-day deployment period.

% \begin{figure*}[t]
%   \centering
%   \includegraphics[width=1\linewidth]{samples/Fig/NJ 24h.jpg}
%   \caption{The variation in PM$_{2.5}$ concentration measurements within the selected sub-region of Nanjing over the course of a day is displayed every 4 hours.}
%   \label{fig:nanjing_24h}
% \end{figure*}

\begin{figure*}[ht]
    \centering
    \subfigure[Nanjing.]{
        \centering
        \includegraphics[width=2.1\columnwidth]{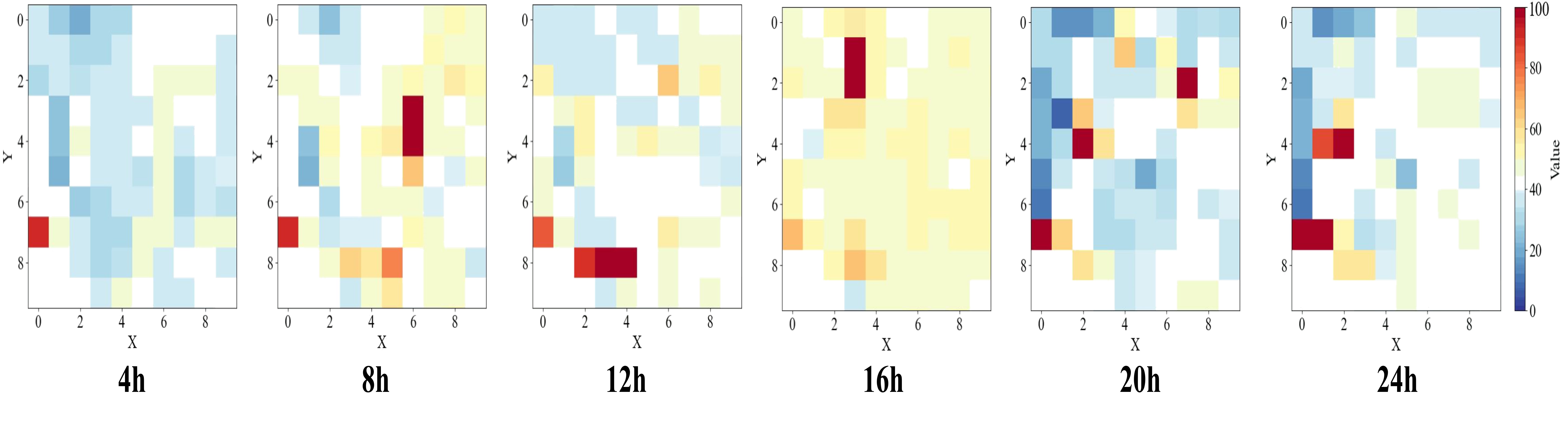}
    }
    \subfigure[Changshu.]{
        \centering
        \includegraphics[width=2.1\columnwidth]{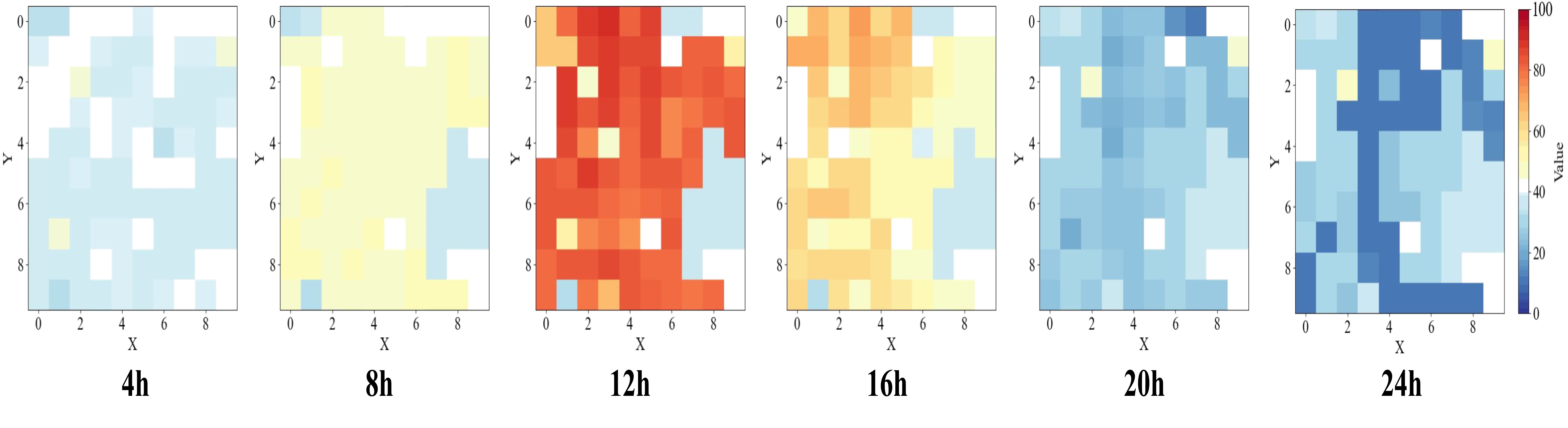}
    }
    \caption{The variation in PM$_{2.5}$ concentration measurements within the selected sub-region of cities over the course of a day is displayed every 4 hours.}
    \label{fig:nanjing_24h}
\end{figure*}

\paragraph{System Parameters}
The spatial resolution is set to 500m$\times$500m, and the temporal resolution is set to 1 hour. Therefore, the air pollution data for a 5km$\times$5km sub-region over the 14-day period can be uniformly transformed into a $\{336, 10, 10\}$ format, where $\{336 = 14 \times 24 \}$ represents the number of time slices, and $\{10, 10\}$ corresponds to the spatial locations at each time slice. Taking Nanjing and Changshu as examples, the spatio-temporal variation of air pollution measurements within a single day is shown in Fig. \ref{fig:nanjing_24h}. The blank areas in the background represent grid cells where no devices passed, resulting in missing measurements. The color gradient, from blue to yellow to red, indicates that the pollutant concentration increases from 0 to 100 $\mu g/m^3$. It is evident that the spatial coverage of measurements dynamically changes over time, and the concentration values for each grid cell also fluctuate. This dynamic and incomplete data poses challenges for our prediction task.
Compared to Changshu, Nanjing has a lower spatial coverage, and the concentration values at each time step exhibit a wider range of variation.
In the spatio-temporal field prediction task, the dataset is generated from the raw spatio-temporal data using a sliding window method, with the implementation details provided in Appendix \ref{A1}.

% As shown in Fig. \ref{fig:slide}, the original spatio-temporal field is divided along the time dimension (the first dimension) into training, validation, and test sets with a 5:1:1 ratio. A sliding window of size "Inputs + Targets" is used to generate the samples. All experiments use the same setting of "12+12," which means utilizing the past 12 time steps to predict the subsequent 12 time steps. Therefore, the measurement data for each selected spatio-temporal region in the city is divided into training sets of size $\{217, 24, 10, 10\}$, validation sets of size $\{25, 24, 10, 10\}$, and test sets of size $\{25, 24, 10, 10\}$, ensuring no data leakage. 
% We train STeP-Diff on the training and validation sets and evaluate the model's performance on the test set.

% \begin{figure*}[t]
%   \centering
%   \includegraphics[width=0.8\linewidth]{samples/Fig/window.jpg}
%   \caption{Taking the PM$_{2.5}$ concentration measurements from Changshu as an example, the dataset is divided using a sliding window. The red line in the figure represents the variation in the average value of each time slice over time. Each sliding window includes 12 time steps as input and 12 time steps as the target prediction values.}
%   \label{fig:slide}
% \end{figure*}

\paragraph{Performance Metric}
% To comprehensively evaluate the prediction performance of models, we use three classic metrics: Mean Absolute Error (MAE), Root Mean Squared Error (RMSE), and Mean Absolute Percentage Error (MAPE).

% MAE measures the average difference between predicted and true values and is not affected by outliers, such as extreme pollution events.
% RMSE reflects the standard deviation of prediction errors and penalizes larger errors more severely. It provides insights into the model's robustness and stability, particularly when predicting spatio-temporal fields where prediction errors for pollution levels can vary significantly.
% MAPE is a relative error metric that expresses the difference between predicted and actual values as a percentage, offering an easily interpretable measure. It is especially valuable for assessing relative prediction errors across different pollution scenarios.

% To comprehensively evaluate model performance, we adopt three metrics: Mean Absolute Error (MAE) measures the average deviation between predicted and true values and is robust to outliers; Root Mean Squared Error (RMSE) places greater emphasis on large errors, reflecting the model’s robustness and stability; Mean Absolute Percentage Error (MAPE) expresses relative errors as percentages, enabling intuitive comparisons across different pollution scenarios.
% The formulas for these metrics are as follows:
To evaluate model performance, we adopt three metrics: Mean Absolute Error (MAE) measures the average deviation between predicted and true values and is robust to outliers; Root Mean Squared Error (RMSE) places greater emphasis on large errors, reflecting the model's robustness and stability; Mean Absolute Percentage Error (MAPE) expresses relative errors as percentages, enabling intuitive comparisons across different pollution scenarios.
The formulas for these metrics are as follows:
\begin{equation}
\begin{aligned}
& \text{MAE} = \frac{1}{N} \sum_{l,x,y} \left| v^{l,x,y} - \hat{v}^{l,x,y} \right|, \\
& \text{RMSE} = \sqrt{\frac{1}{N} \sum_{l,x,y} \left( v^{l,x,y} - \hat{v}^{l,x,y} \right)^2}, \\
& 
\text{MAPE} = \frac{1}{N} \sum_{l,x,y} \left| \frac{v^{l,x,y} - \hat{v}^{l,x,y}}{v^{l,x,y}} \right|,
\end{aligned}
\end{equation}
where $v^{l,x,y}$ represents the true value at the $(x,y)$ position at time $l$, and $\hat{v}^{l,x,y}$ denotes the predicted value at the same $(x,y)$ position and time $l$, $N$ is total number of data points.

\paragraph{Baselines}
% We elaborately select the following spatial-temporal forecasting algorithms to be compared with our proposed model, which covers the most representative methods proposed in recent years.
% {\color{blue}
We selected representative models from three main categories for comparative analysis with our proposed approach: PDE as a physics-driven method; NeuralODE, GinAR, UniST, DDPM, DeepONet, CSDI, STGNN and AirFormer as data-driven methods; and HMSS, AirPhyNet, Phy-APMR, DiffusionPDE and PINN-Diffusion as hybrid models integrating physical principles with deep learning.
% }

\begin{itemize}
    % {\color{blue}
    \item \textit{PDE} describes the physical principles of convection and diffusion in air pollutant dispersion \cite{zannetti2013air}.
    % }
    \item \textit{NeuralODE} uses neural networks to model hidden state dynamics for capturing pollution patterns \cite{Chen2018}.
    \item \textit{GinAR}
    uses an attention-based recurrent graph network to model spatio-temporal dependencies \cite{Yu2024}.
    \item \textit{UniST} designs a universal large-scale language model for general urban spatio-temporal forecasting \cite{Yuan2024}.
    \item \textit{DDPM} employs a diffusion probabilistic model to generate predictions by adding noise to target locations \cite{Ho2020}.  
    \item \textit{DeepONet} learns spatiotemporal patterns from distributed data for accurate value prediction\cite{Lu2021}.
    \item \textit{CSDI} utilizes a conditional score-based diffusion model to capture correlations for time-series forecasting \cite{Tashiro2021}.
     % {\color{blue}
    \item \textit{STGNN} captures multi-scale spatiotemporal dependencies through graph convolutions, enabling efficient and accurate time-series forecasting \cite{yu2017spatio}.
    \item \textit{AirFormer} uses a two-stage Transformer combining efficient spatiotemporal self-attention and latent variable modeling to greatly enhance prediction accuracy \cite{liang2023airformer}.
    % }
    \item \textit{HMSS} integrates physics and data-driven models for robust fine-grained predictions \cite{Chen2020}.
    % {\color{blue}
    \item \textit{Phy-AMPR} combines an inference network with a physics-informed network encoding pollutant transport PDEs to jointly reduce data and physical errors \cite{shi2025phy}.
    % {\color{blue}
    \item \textit{DiffusionPDE} incorporates PDE structural information effectively during the diffusion sampling process by jointly modeling the solution and coefficient spaces \cite{huang2024diffusionpde}.
    \item \textit{PINN-Diffusion} leverages known partial differential equations as physical priors for conditional generation, demonstrating strong performance under irregular or sparse observations \cite{shu2023physics}.
    \item \textit{AirPhyNet} uses Neural ODEs with graph-based representations to model pollutant diffusion and advection \cite{hettige2024airphynet}.
    \item \textit{Air-DualODE} 
    employs a dual-branch Neural ODE to simulate physical processes and learn environmental dependencies, enhancing accuracy through fusion~\cite{tian2024air}.
    % uses a dual-branch Neural ODE: one for physical simulation of diffusion-advection, another for spatiotemporal dependencies of environmental factors, fused to improve accuracy.
    \item \textit{STPP} 
    learns spatiotemporal joint distributions via a diffusion model and captures event correlations with a co-attention module~\cite{yuan2023spatio}.
    % adopts a diffusion model framework to learn complex spatiotemporal joint distributions via multi-step Gaussian diffusion and adaptively captures event dependencies with a spatiotemporal co-attention module. 
\end{itemize}

\begin{table*}[t]
\caption{
Performance on our datasets, where bold denotes the best results and underline denotes the second best results. Mobile refers to the measurements from mobile devices as the ground truth, while National refers to the measurements from national control stations as the ground truth.
% Symbol (*) indicates statistical significance ($p < 0.05$) using a paired t-test compared to the suboptimal model.
% Statistical significance levels are denoted as follows:
% ***$ p < 0.001$; **$ 0.001 < p < 0.01$; *$ 0.01 < p < 0.05$.
The symbol (*) indicates statistical significance compared to the suboptimal model based on a paired t-test, with significance levels defined as: *** for $p < 0.001$, ** for $0.001 < p < 0.01$, and * for $0.01 < p < 0.05$.
}
\label{SOTA}
\centering
\normalsize
% \footnotesize  % 比 \normalsize 更紧凑，避免换行
% \begin{tabularx}{\textwidth}{l *{12}{>{\centering\arraybackslash}X}}
% \begin{tabularx}{\textwidth}{l*{12}{>{\centering\arraybackslash}p{1.05 cm}}}
\renewcommand{\arraystretch}{0.95}
\begin{tabularx}{\textwidth}{l@{\hskip 1pt}*{12}{>{\centering\arraybackslash}p{1.3cm}@{\hskip 0.0pt}}}
\toprule
\textbf{City} & \multicolumn{3}{c}{\textbf{Changshu (Mobile)}} & \multicolumn{3}{c}{\textbf{Nanjing (Mobile)}} & \multicolumn{3}{c}{\textbf{Changshu (National)}} & \multicolumn{3}{c}{\textbf{Nanjing (National)}} \\
\cmidrule(lr){2-4} \cmidrule(lr){5-7} \cmidrule(lr){8-10} \cmidrule(lr){11-13}
\textbf{Metrics} & MAE & RMSE & MAPE & MAE & RMSE & MAPE & MAE & RMSE & MAPE & MAE & RMSE & MAPE \\
\midrule
% \textcolor{blue}{PDE} & \textcolor{blue}{19.65} & \textcolor{blue}{24.80} & \textcolor{blue}{0.48} & \textcolor{blue}{16.42} & \textcolor{blue}{25.51} & \textcolor{blue}{0.77} & \textcolor{blue}{17.03} & \textcolor{blue}{24.14} & \textcolor{blue}{0.56} & \textcolor{blue}{21.53} & \textcolor{blue}{25.91} & \textcolor{blue}{0.83} \\
PDE & 19.65 & 24.80 & 0.48 & 16.42 & 25.51 & 0.77 & 17.03 & 24.14 & 0.56 & 21.53 & 25.91 & 0.83 \\
\midrule
NeuralODE & 30.24 & 34.10 & 0.73 & 19.15 & 26.91 & 0.92 & 33.30 & 40.00 & 1.03 & 21.09 & 23.42 & 0.90 \\
GinAR & 17.29 & 22.06 & 0.43 & 21.53 & 29.58 & 1.77 & \underline{8.07} & \underline{9.60} & 0.41 & 7.70 & 10.16 & 0.77 \\
UniST & 8.48 & \underline{12.35} & \underline{0.28} & 25.13 & 33.22 & 0.59 & 14.82 & 17.00 & 0.47 & 43.45 & 44.34 & 1.74 \\
DDPM & 37.21 & 37.69 & 0.99 & 27.94 & 34.70 & 0.97 & 39.81 & 41.09 & 0.89 & 25.27 & 27.01 & 0.96 \\
DeepONet & 10.41 & 13.47 & 0.30 & 15.69 & 24.27 & \underline{0.52} & 8.97 & 12.85 & \underline{0.28} & 11.97 & 14.51 & 0.77 \\
CSDI & \underline{8.09} & 16.47 & 1.60 & \underline{3.34} & \underline{8.80} & 0.56 & 8.95 & 18.45 & 0.99 & \underline{2.50} & \underline{7.44} & \textbf{0.33} \\
% \textcolor{blue}{STGNN} & \textcolor{blue}{26.36} & \textcolor{blue}{28.61} & \textcolor{blue}{0.55} & \textcolor{blue}{9.97} & \textcolor{blue}{17.49} & \textcolor{blue}{0.43} & \textcolor{blue}{40.05} & \textcolor{blue}{41.17} & \textcolor{blue}{0.98} & \textcolor{blue}{23.78} & \textcolor{blue}{24.46} & \textcolor{blue}{0.89} \\
% \textcolor{blue}{AirFormer} & \textcolor{blue}{24.29} & \textcolor{blue}{27.04} & \textcolor{blue}{0.51} & \textcolor{blue}{10.01} & \textcolor{blue}{16.93} & \textcolor{blue}{0.52} & \textcolor{blue}{8.75} & \textcolor{blue}{10.89} & \textcolor{blue}{0.54} & \textcolor{blue}{7.88} & \textcolor{blue}{9.71} & \textcolor{blue}{0.46} \\
STGNN & 26.36 & 28.61 & 0.55 & 9.97 & 17.49 & 0.43 & 40.05 & 41.17 & 0.98 & 23.78 & 24.46 & 0.89 \\
AirFormer & 24.29 & 27.04 & 0.51 & 10.01 & 16.93 & 0.52 & 8.75 & 10.89 & 0.54 & 7.88 & 9.71 & 0.46 \\
\midrule
HMSS & 16.40 & 15.29 & 0.36 & 15.69 & 24.27 & 0.52 & 18.50 & 17.50 & 0.46 & 11.97 & 14.51 & 0.71 \\
Phy-AMPR & 12.58 & 12.51 & 0.31 & 13.02 & 19.78 & 0.61 & 15.31 & 14.52 & 0.37 & 9.84 & 11.87 & 0.53 \\
DiffusionPDE & 8.77 & 12.75 & 0.29 & 10.20 & 15.97 & 0.54 & 8.81 & 9.61 & 0.29 & 7.78 & 9.43 & 0.50 \\
PINN-Diffusion & 13.11 & 13.51 & 0.38 & 14.59 & 10.88 & 0.58 & 17.87 & 12.88 & 0.41 & 10.22 & 11.61 & 0.37 \\
% \color{blue}{AirPhyNet} & \textcolor{blue}{11.14} & \textcolor{blue}{13.43} & \textcolor{blue}{0.30} & \textcolor{blue}{9.39} & \textcolor{blue}{14.86} & \textcolor{blue}{0.79} & \textcolor{blue}{11.53} & \textcolor{blue}{12.84} & \textcolor{blue}{0.44} & \textcolor{blue}{10.23} & \textcolor{blue}{12.11} & \textcolor{blue}{0.42} \\
AirPhyNet & 11.14 & 13.43 & 0.30 & 9.39 & 14.86 & 0.79 & 11.53 & 12.84 & 0.44 & 10.23 & 12.11 & 0.42 \\
% \textcolor{blue}{Air-DualODE} & \textcolor{blue}{10.52} & \textcolor{blue}{12.89} & \textcolor{blue}{0.29} & \textcolor{blue}{8.83} & \textcolor{blue}{14.16} & \textcolor{blue}{0.74} & \textcolor{blue}{11.07} & \textcolor{blue}{12.20} & \textcolor{blue}{0.42} & \textcolor{blue}{9.73} & \textcolor{blue}{11.38} & \textcolor{blue}{0.44} \\
% \textcolor{blue}{STPP} & \textcolor{blue}{10.62} & \textcolor{blue}{15.13} & \textcolor{blue}{0.40} & \textcolor{blue}{8.87} & \textcolor{blue}{15.62} & \textcolor{blue}{0.69} & \textcolor{blue}{9.20} & \textcolor{blue}{14.73} & \textcolor{blue}{0.46} & \textcolor{blue}{11.63} & \textcolor{blue}{15.81} & \textcolor{blue}{0.69} \\
Air-DualODE & 10.52 & 12.89 & 0.29 & 8.83 & 14.16 & 0.74 & 11.07 & 12.20 & 0.42 & 9.73 & 11.38 & 0.44 \\
STPP & 10.62 & 15.13 & 0.40 & 8.87 & 15.62 & 0.69 & 9.20 & 14.73 & 0.46 & 11.63 & 15.81 & 0.69 \\
\midrule
% \textbf{STeP-Diff} & \textbf{1.30}\textcolor{blue}{***} & \textbf{3.52}\textcolor{blue}{***} & \textbf{0.17}\textcolor{blue}{*} & \textbf{0.95}\textcolor{blue}{***} & \textbf{2.94}\textcolor{blue}{***} & \textbf{0.15}\textcolor{blue}{*} & \textbf{0.88}\textcolor{blue}{***} & \textbf{2.40}\textcolor{blue}{**} & \textbf{0.25}\textcolor{blue}{*} & \textbf{0.84}\textcolor{blue}{***} & \textbf{3.25}\textcolor{blue}{***} & \underline{0.41} \\
\textbf{STeP-Diff} & \textbf{1.30}*** & \textbf{3.52}*** & \textbf{0.17}* & \textbf{0.95}*** & \textbf{2.94}*** & \textbf{0.15}* & \textbf{0.88}*** & \textbf{2.40}** & \textbf{0.25}* & \textbf{0.84}*** & \textbf{3.25}*** & \underline{0.41} \\
\bottomrule
\end{tabularx}
\end{table*}

\begin{figure*}[ht]
    \centering
    \subfigure[Nanjing.]{
        \centering
        \includegraphics[width=2.1\columnwidth]{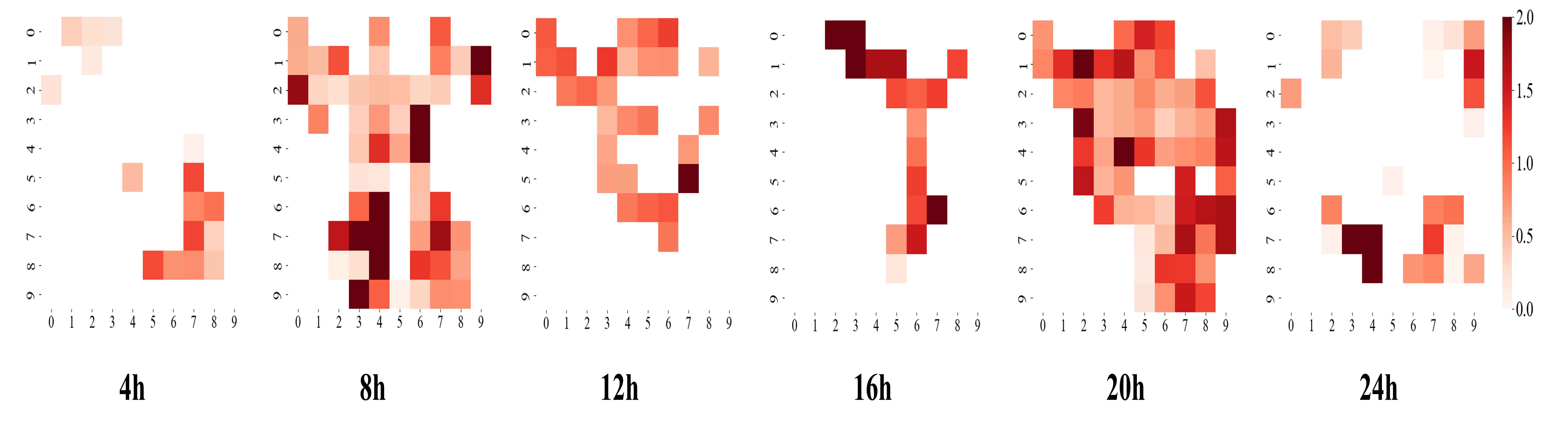}
    }
    \subfigure[Changshu.]{
        \centering
        \includegraphics[width=2.1\columnwidth]{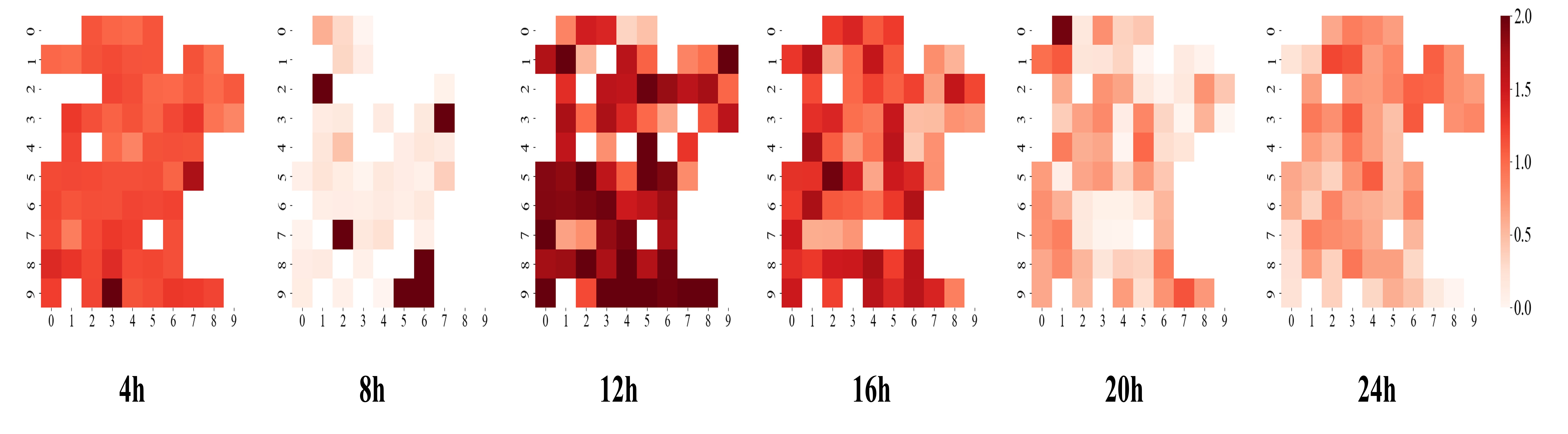}
    }
    \caption{Visualization of model reconstruction error over a day in selected subregions.}
    \label{error}
\end{figure*}

% \subsection{System Performance}
\subsection{Experimental Results}

\label{Sys}

In Section \ref{4.3.1}, we evaluate the overall performance metrics of our proposed method in comparison with state-of-the-art (SOTA) approaches. Section \ref{4.3.2} presents the visualization of the reconstruction results. In Section \ref{4.3.3}, we conduct a stratified performance analysis of the experimental results, focusing on different time periods and spatial locations. Section \ref{4.3.4} investigates the impact of various parameter settings on model performance. 
% In Section \ref{4.3.5}, we perform ablation studies to assess the contributions of the two core components of our model. Finally, Section \ref{4.3.6} explores the effects of different physics-informed integration strategies on model performance.
Finally, in Section \ref{4.3.5}, we perform ablation studies to assess the contributions of the two core components of our model. Additionally, we investigate the effects of different physics-informed integration strategies on model performance in Appendix \ref{A3}.

\subsubsection{Overall Performance}
\label{4.3.1}
We evaluated the performance of the proposed model against several baselines on datasets from two cities, considering two settings where either mobile device measurements or national monitoring station data were used as ground truth. As shown in Table \ref{SOTA}, our model achieved the best results across all four datasets, significantly outperforming existing methods. Compared to the second-best models, our approach improved MAE, RMSE, and MAPE by 89.12\%, 82.30\%, and 25.00\% on Changshu (Mobile); 76.65\%, 66.02\%, and 13.46\% on Nanjing (Mobile); 89.10\%, 75.00\%, and 10.71\% on Changshu (National); and 66.40\%, 56.32\%, and -24.23\% on Nanjing (National), respectively.
  
Further comparison reveals that when mobile data serve as ground truth, the model performs better on the Nanjing dataset than on Changshu’s, despite Changshu having greater spatiotemporal coverage. This is likely due to the larger number of evaluation points in Changshu, which increases prediction difficulty. Conversely, when national station data are used as ground truth, the model achieves better performance on Changshu, likely because of its fewer national stations, resulting in lower prediction complexity. Moreover, most state-of-the-art methods perform better with mobile data as ground truth, as training data comprise solely mobile sensor measurements and mobile trajectories near national stations tend to be sparse, limiting prediction accuracy. In contrast, our model demonstrates superior performance even near national stations, indicating its stronger capability to capture true air pollution dynamics and greater robustness to measurement noise.

In addition, we conducted statistical significance testing on 12 evaluation metrics by comparing our model with the second-best baseline. For each metric, we performed 10 independent runs and applied a paired t-test for analysis. The results show that 11 out of 12 metrics yield p-values smaller than 0.05, indicating that our model consistently provides statistically significant improvements over the suboptimal model. Notably, 7 of these comparisons achieve p-values below 0.001, further highlighting the substantial and stable advantages of our method across multiple dimensions.The detailed p-values are presented in Table~\ref{p-value}.

Furthermore, we present a systematic analysis in Appendix~\ref{time} of the training and inference time of STeP-Diff and competing methods, evaluated on a single NVIDIA RTX A6000 GPU. The results demonstrate that STeP-Diff achieves a favorable balance between training efficiency and model performance, making it particularly suitable for offline forecasting tasks and applications requiring high prediction accuracy.

Finally, we provide a supplementary decision-oriented evaluation of the model under pollution warning scenarios in Appendix~\ref{warn}, aiming to comprehensively assess the model’s ability to identify pollution days and the overall performance of its warnings. The experimental results further validate the robustness and practical utility of STeP-Diff in real-world pollution warning applications, highlighting its broad potential for air quality management.

% }

\subsubsection{Reconstruction Visualization}
\label{4.3.2}

\begin{figure*}[ht]
  \centering
  \subfigure[Performance vs. Time.]{
    \centering
    \includegraphics[height=0.49\columnwidth]{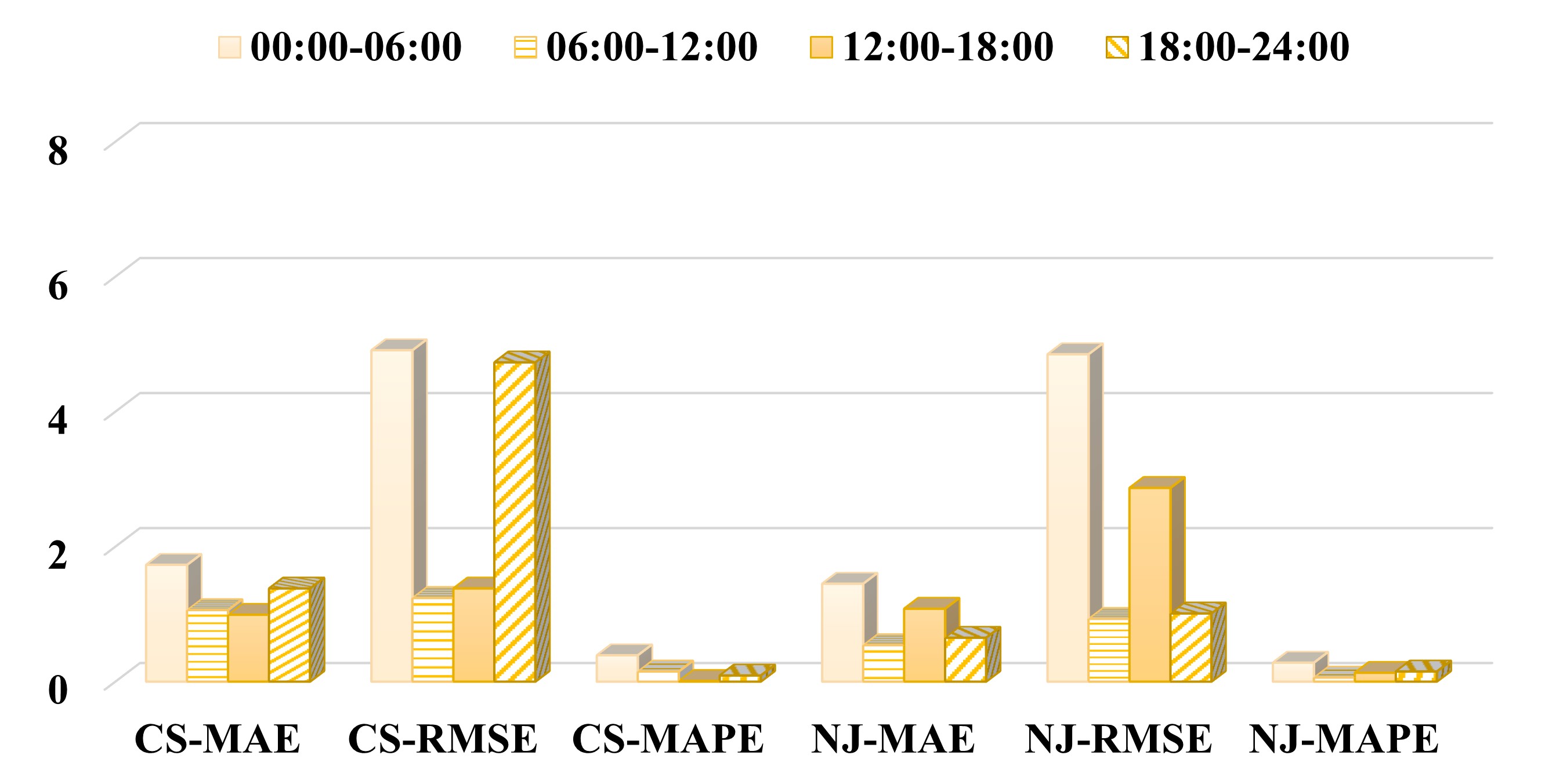}
  }
  \subfigure[Performance vs. Locations.]{
    \centering
    \includegraphics[height=0.49\columnwidth]{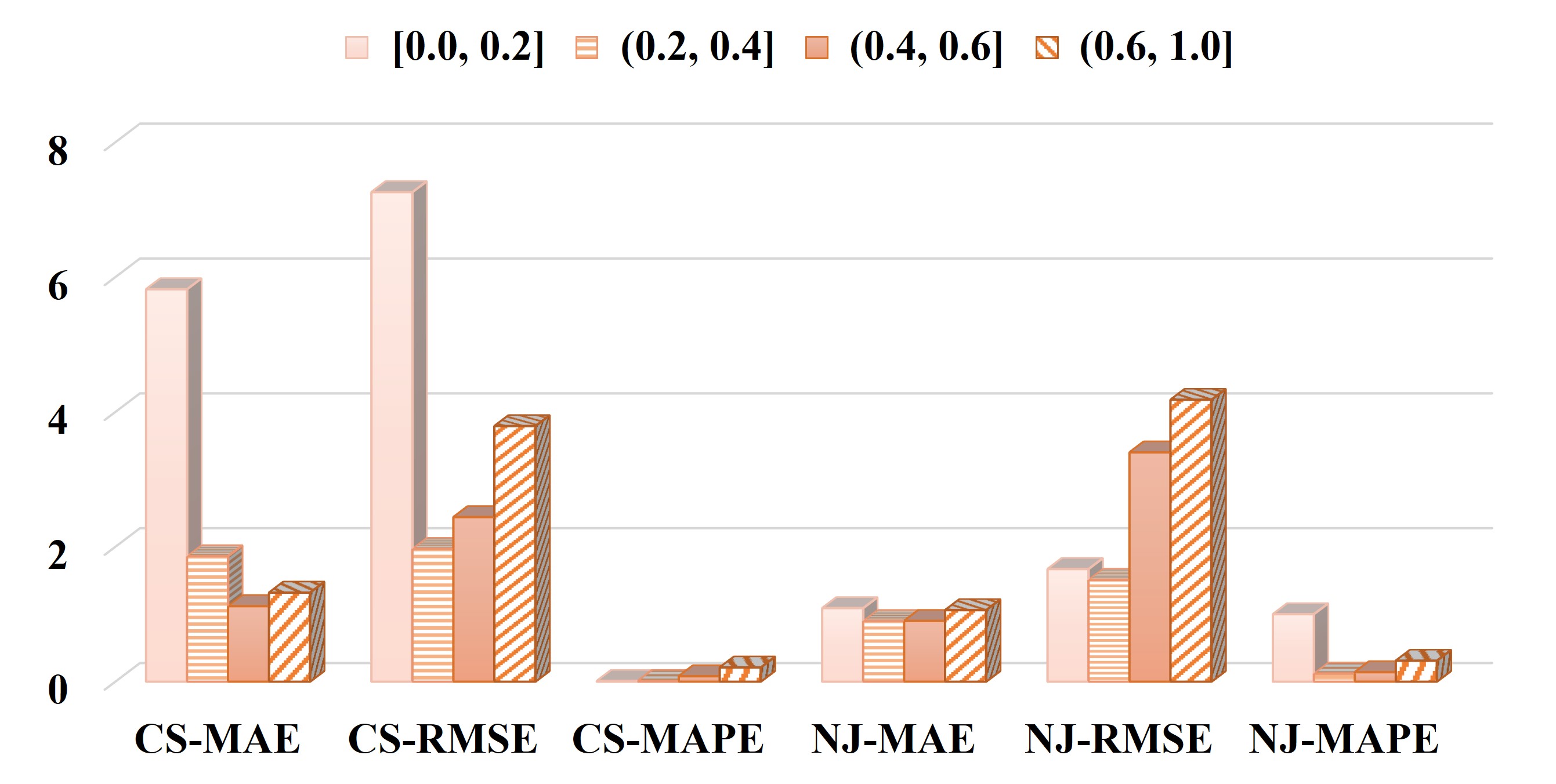}
  }
  \caption{Impact of forecasting time and locations on model performance. CS represents Changshu, and NJ represents Nanjing. We compared the performance of three metrics (MAE, MSE, MAPE) for the two cities using mobile device measurements as ground truth, under four time periods within a day and four spatial coverage intervals.}
  \label{vs}
\end{figure*}

\begin{figure*}[t]
  \centering
  \subfigure[Impact of \(\omega\).]{
    \centering
    \includegraphics[height=0.46\columnwidth]{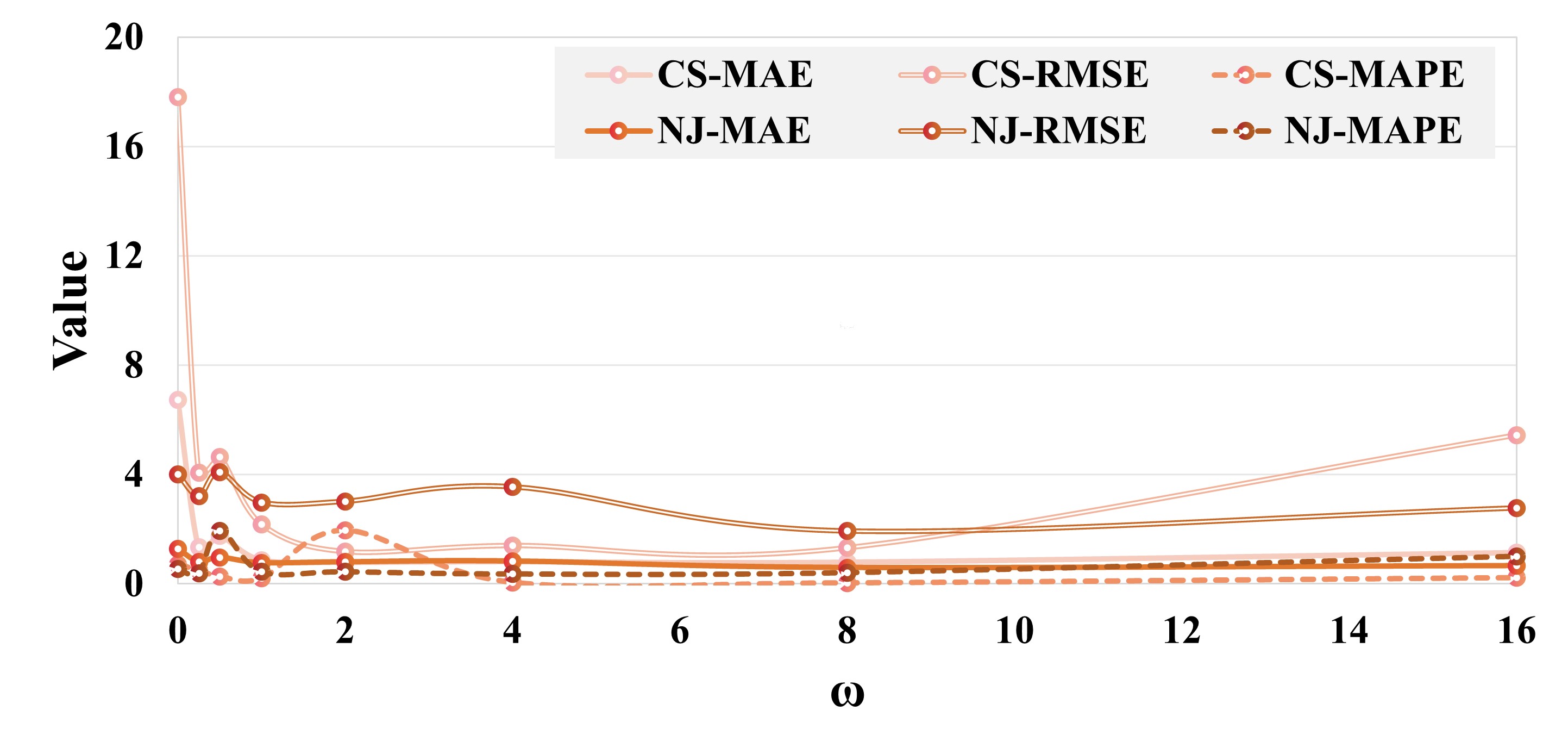}
  }
  \subfigure[Spatio-temporal Transformer layers.]{
    \centering
    \includegraphics[height=0.46\columnwidth]{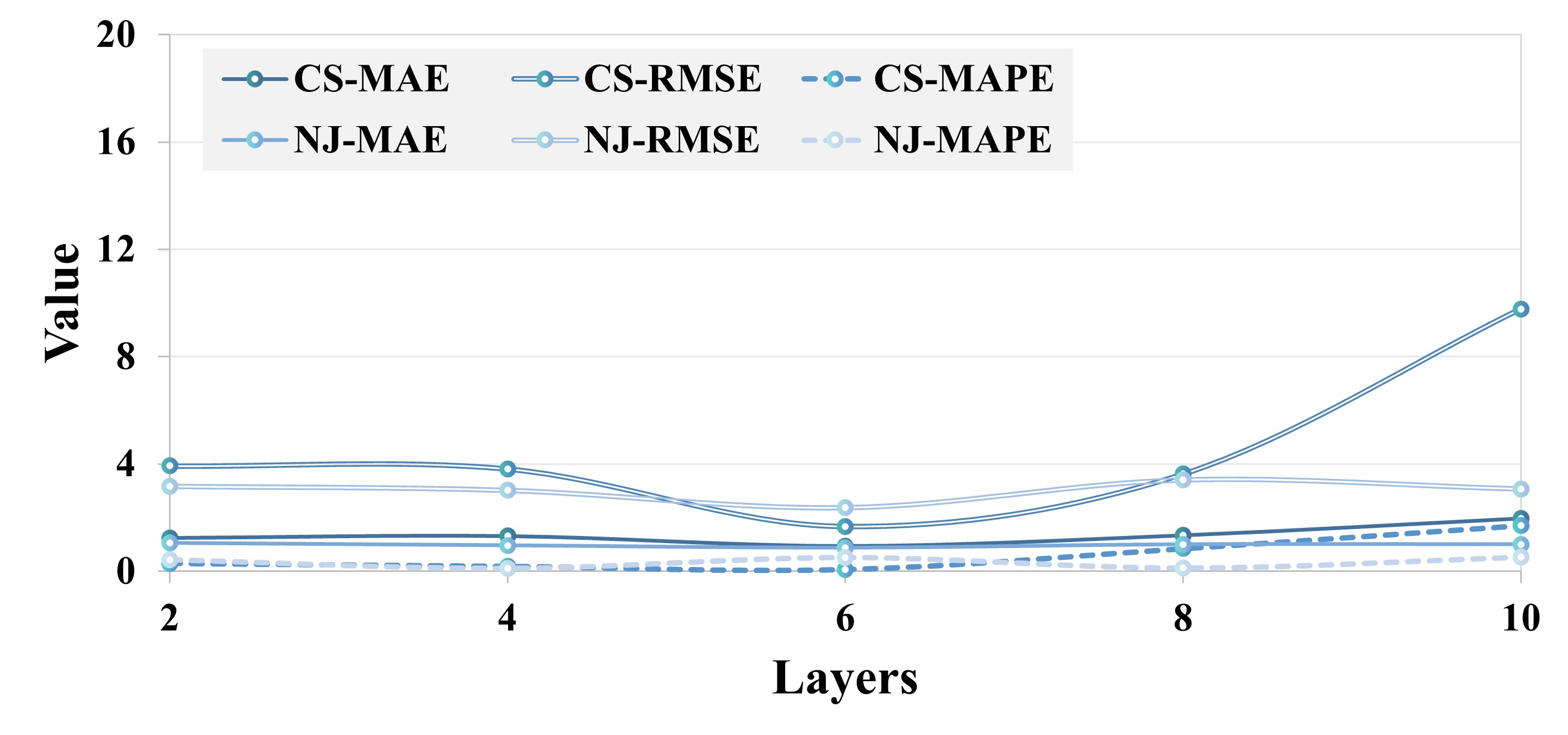}
  }
  \caption{Impact of parameters on model performance.}
  \label{fig:parameter}
\end{figure*}

\begin{table*}[t]
\caption{Performance of ablation experiments, where bold denotes the best results and underline denotes the second best results.}
\label{Abaltion}
\centering
\normalsize
\setlength{\tabcolsep}{5pt} % 调整列间距
\renewcommand{\arraystretch}{0.95}
\begin{tabularx}{\textwidth}{l*{12}{>{\centering\arraybackslash}p{0.85 cm}}}
\toprule
\textbf{City} & \multicolumn{3}{c}{\textbf{Changshu (Mobile)}} & \multicolumn{3}{c}{\textbf{Nanjing (Mobile)}} & \multicolumn{3}{c}{\textbf{Changshu (National)}} & \multicolumn{3}{c}{\textbf{Nanjing (National)}} \\
\cmidrule(lr){2-4} \cmidrule(lr){5-7} \cmidrule(lr){8-10} \cmidrule(lr){11-13}
\textbf{Metrics} & MAE & RMSE & MAPE & MAE & RMSE & MAPE & MAE & RMSE & MAPE & MAE & RMSE & MAPE \\
\midrule
STeP-Diff$-$PDE & 6.74 & 17.81 & 0.72 & 1.30 & 4.01 & 0.55 & 6.64 & 17.58 & 0.57 & 1.29 & 3.97 & 0.35 \\
STeP-Diff$-$DeepONet & \underline{2.25} & \underline{11.56} & \underline{1.18} & \underline{1.64} & \underline{4.61} & \underline{0.43} & \underline{3.50} & \underline{12.90} & \underline{0.93} & \underline{1.38} & \underline{3.88} & \underline{0.73} \\
\textbf{STeP-Diff} & \textbf{1.30} & \textbf{3.52}& \textbf{0.17} & \textbf{0.95} & \textbf{2.94} & \textbf{0.15} & \textbf{0.88} & \textbf{2.40} & \textbf{0.25} & \textbf{0.84} & \textbf{3.25} & \textbf{0.41} \\
\bottomrule
\end{tabularx}
\end{table*}

To provide a more intuitive visualization of the reconstruction performance, we selected six time points (3h, 7h, 11h, 15h, 19h, and 23h) from the test sets of two cities to illustrate the reconstruction error. As shown in Fig. \ref{error}, the error values range from 0 to 2 $\mu g/m^3$, where increasing error values are represented by a deepening red color, while white areas indicate the absence of measurements. The spatial coverage varies across different time slots, influenced by both the operational frequency of mobile sensing devices and urban mobility patterns. Overall, the coverage in Changshu is higher than in Nanjing, as buses in Changshu provide more extensive and regular spatial coverage compared to cars in Nanjing. The prediction error also fluctuates over time, driven by variations in pollutant concentrations and the spatio-temporal coverage of surrounding measurements. Although Changshu exhibits higher spatio-temporal coverage, its overall error values are higher than those in Nanjing. This is primarily due to the greater number of evaluation locations in Changshu, which makes reducing the error metrics more challenging. However, this does not imply that the prediction performance in Changshu is inferior to that in Nanjing.

\subsubsection{Stratified Performance Analysis}
\label{4.3.3}

In this section, we analyze the impact of different time periods and spatial coverage intervals on the model's performance.

\paragraph{Performance vs. Time}
To evaluate the prediction accuracy of our method across different time periods, we calculate the metric for four time periods (00:00-06:00, 06:00-12:00, 12:00-18:00, and 18:00-24:00) using the measurement data collected from mobile devices in two cities, as shown in Fig. \ref{vs}(a). Specifically, for the pollution dataset collected in Changshu, the performance is the worst from 00:00 to 06:00, primarily due to the near cessation of bus operations during this period, resulting in the lowest spatio-temporal coverage and insufficient exploitation of spatio-temporal features. The performance is better from 06:00 to 12:00, as the number of bus trips gradually increases, leading to higher spatio-temporal coverage. The performance is the best from 12:00 to 18:00, when the bus operation rate peaks and the coverage reaches its maximum. The performance is relatively poor from 18:00 to 24:00, as the number of bus trips gradually decreases, resulting in lower coverage.
In comparison, for the pollution dataset collected in Nanjing, 
% the performance is the worst from 00:00 to 06:00, due to the lowest demand for car rides and the lowest spatio-temporal coverage during this period. The performance is the best from 06:00 to 12:00, which includes the morning rush hour, when the demand for car rides increases, resulting in higher coverage. The performance is poorer from 12:00 to 18:00, as the demand for rides gradually decreases, leading to lower coverage. The performance improves from 18:00 to 24:00, as this period includes the evening rush hour, when the demand for rides increases again, resulting in higher coverage.
the performance is the worst from 00:00 to 06:00, as the vehicle operating rate is the lowest during this period, resulting in the weakest spatio-temporal coverage. The performance is the best from 06:00 to 12:00, as this period corresponds to the main operational hours for vehicles, with increased demand for services leading to higher spatio-temporal coverage. From 12:00 to 18:00, the performance declines due to a gradual decrease in demand, which reduces the spatio-temporal coverage. The performance improves from 18:00 to 24:00, as human activity increases during this period, leading to a resurgence in operational demand and consequently higher spatio-temporal coverage.
 
% \begin{table*}[t]
% \caption{Performance Metrics for Different Forecasting Time}
% \label{Time}
% \centering
% \scriptsize
% \setlength{\tabcolsep}{4pt} % 设置列间距，可以根据需要调整
% \begin{tabular}{c*{6}{>{\centering\arraybackslash}p{1.0cm}}}
% \toprule
% \textbf{City} & \multicolumn{3}{c}{\textbf{Changshu (Mobile)}} & \multicolumn{3}{c}{\textbf{Nanjing (Mobile)}} \\
% \cmidrule(lr){2-4} \cmidrule(lr){5-7}
% \textbf{Metric} & \textbf{MAE} & \textbf{RMSE} & \textbf{MAPE} & \textbf{MAE} & \textbf{RMSE} & \textbf{MAPE} \\
% \midrule
% \textbf{00:00--06:00} & 1.73 & 4.91 & 0.39 & 1.45 & 4.85 & 0.28 \\
% \textbf{06:00--12:00} & 1.06 & 1.24 & 0.15 & 0.54 & 0.93 & 0.06 \\
% \textbf{12:00--18:00} & 0.99 & 1.38 & 0.01 & 1.08 & 2.87 & 0.13 \\
% \textbf{18:00--24:00} & 1.38 & 4.73 & 0.09 & 0.65 & 1.01 & 0.15 \\
% \bottomrule
% \end{tabular}
% \end{table*}

\paragraph{Performance vs. Location}
To assess the prediction accuracy of our method in regions with different coverage levels, we computed performance metrics based on mobile measurement datasets from two cities, categorized into coverage intervals [0.0, 0.2], (0.2, 0.4], (0.4, 0.6], and (0.6, 1.0]. The results are shown in Fig. \ref{vs}(b), with the corresponding spatial distributions visualized in different green-shaded areas areas on the left side of Fig. \ref{fig:space}(a)(b).
In the Changshu dataset, regions with coverage [0.0, 0.2] performed the worst, as these areas were predominantly villages with dense pollution sources and low spatio-temporal coverage, resulting in poor prediction accuracy. 
Regions with coverage (0.2, 0.4] showed improvement, mainly consisting of villages, where the increase in spatio-temporal coverage somewhat optimized the prediction performance. 
The highest prediction accuracy was observed in regions with coverage (0.4, 0.6], primarily residential communities with convenient transportation and higher spatio-temporal coverage, leading to optimal prediction performance. 
Areas with coverage (0.6, 1.0] performed well, which mainly consisted of tourist spots and hotels, with few pollution sources and the highest spatio-temporal coverage, which improved prediction accuracy.
% In the Nanjing dataset, regions with coverage [0.0, 0.2] exhibited the lowest prediction accuracy (ranked fourth), primarily consisting of building areas with numerous pollution sources and low coverage. 
% Areas with coverage (0.2, 0.4] performed better (ranked second), primarily villages with fewer pollution sources and moderate spatio-temporal coverage, improving prediction performance. 
% Regions with coverage (0.4, 0.6] showed lower prediction accuracy (ranked third), still consisting mainly of villages, but the high surrounding traffic flow and fluctuating pollutant concentrations made spatio-temporal feature extraction more difficult. 
% The best performance (ranked first) was observed in areas with coverage (0.6, 1.0], primarily consisting of major roads with few pollution sources and the highest spatio-temporal coverage, resulting in the highest prediction accuracy.
In the Nanjing dataset, regions with coverage [0.0, 0.2] performed relatively well, as these areas had the lowest coverage but fewer pollution sources, resulting in lower prediction difficulty.
Regions with coverage (0.2, 0.4] performed the best, primarily consisting of major roads with relatively stable pollutant concentrations.
Regions with coverage (0.4, 0.6] exhibited lower prediction accuracy, mainly consisting of building areas with numerous pollution sources, leading to higher prediction difficulty.
Regions with coverage (0.6, 1.0] performed the worst, despite having the highest coverage, as these areas were predominantly villages, and the high surrounding traffic flow and fluctuating pollutant concentrations made spatio-temporal feature extraction more challenging.

\subsubsection{Parameters Analysis}
\label{4.3.4}

In this section, we analyze the impact of the 
$\omega$ weight in the loss function and the number of layers of the Transformer in the denoising network on the model's performance.

\paragraph{Impact of \(\omega\)}
To further evaluate the impact of the weight \(\omega\) in the loss function on model performance (defaulted to 1 in other experiments), 
we plotted the variation of performance metrics for different \(\omega\) values of 0, 0.25, 0.5, 1, 2, 4, 8, and 16 in Fig. \ref{fig:parameter}(a). When \(\omega = 0\), the model performs the worst, which is equivalent to not incorporating the PDE physical constraints. As \(\omega\) increases, the model's performance improves, with the metrics gradually decreasing. The model reaches its best performance when \(\omega = 8\), but further increases in \(\omega\) lead to a deterioration in performance, with the metrics increasing again.
This indicates that \(\omega\) has a nonlinear impact on model performance. A moderate increase in 
\(\omega\) can enhance performance, but exceeding a certain threshold may lead to a decline in performance.

\paragraph{Transformer Layers}
To further assess the impact of the number of spatio-temporal Transformer layers on the performance of the denoising network (defaulted to 4 in other experiments), we plotted the variation of performance metrics for layer counts of 2, 4, 6, 8, and 10 in Fig. \ref{fig:parameter}(b).
% As shown, increasing the number of layers leads to a gradual improvement in model performance (overall reduction in metric values).
% The model achieves optimal performance with six layers, where the metric value is the lowest.
% However, as the number of layers increases beyond six, the model's performance deteriorates (metric values increase).
% This indicates that the number of spatio-temporal Transformer layers also has a nonlinear effect on model performance; an appropriate number of layers can enhance performance, but excessive layers may lead to a decline.
As shown, increasing the number of layers results in a gradual improvement in model performance, as indicated by a consistent reduction in metric values. The model achieves optimal performance with six layers, where the metric value reaches its lowest point. However, as the number of layers exceeds six, the model's performance begins to deteriorate, with an increase in metric values. This suggests that the number of spatio-temporal Transformer layers has a nonlinear effect on model performance. An appropriate number of layers can enhance performance, but excessive layers may lead to a decline.

\subsubsection{Ablation Study}
\label{4.3.5}

We conducted experimental analyses on the core components of our model, focusing on each component's contribution to the overall system performance, as shown in Table \ref{Abaltion}. After incorporating the PDE component, the model's performance improved across four datasets, with MAE, RMSE, and MAPE increasing by 80.71\%, 80.24\%, and 76.30\%; 26.92\%, 26.68\%, and 72.73\% ; 86.75\%, 86.35\%, and 56.14\%; and 34.88\%, 18.13\%, and -17.14\%. In contrast, adding the DeepONet component resulted in MAE, RMSE, and MAPE improvements of 42.22\%, 69.55\%, and 85.60\%; 42.07\%, 79.61\%, and 65.12\%; 74.86\%, 81.40\%, and 73.12\%; and 9.13\%, 16.24\%, and 43.84\%. 
% These results indicate that the physical constraints imposed by the PDE component significantly enhance model performance compared to the spatial information provided by DeepONet. Notably, all three experimental outcomes in the table outperform the current SOTA methods.
These findings confirm that the physical constraints imposed by the PDE component significantly enhance model performance compared to the spatial information provided by DeepONet. All three experimental outcomes outperform the current SOTA methods.

\section{Related work and discussion}

\label{rel}

In this section, we have reviewed and discussed the relevant research in the fields of air pollution modeling (in Section \ref{5.1}) and urban spatio-temporal forecasting (in Section \ref{5.2}).

\subsection{Air Pollution Modeling}
\label{5.1}

Air pollution modeling can be broadly categorized into three main approaches: physics-based methods, data-driven methods, and hybrid physics-data driven methods.

\subsubsection{Physics-based Methods}
Physics-based partial differential equation (PDE) methods have achieved notable progress in air pollution monitoring \cite{Holmes06}, enabling high-resolution predictions through numerical simulations \cite{zannetti2013air}.
However, their performance heavily depends on parameters that are often difficult to obtain and may vary over time, leading to model inaccuracies and reduced prediction reliability.

\subsubsection{Data-driven Methods}
With the advancement of big data technologies, machine learning and deep learning have been widely applied to modeling air pollution dispersion \cite{zhou2025catua}.
For example, U-air employs a co-training approach to infer fine-grained pollution levels across an entire city \cite{zheng2013u}, sparking significant interest in fine-grained air quality inference as a key challenge in smart cities \cite{ma2020fine, liu2024mobiair}.
Models such as SSH-GNN and DAL further leverage semi-supervised learning for air quality prediction and analysis \cite{han2022semi, qi2018deep}.
FineMoGen~\cite{zhang2023finemogen} explicitly models spatiotemporal compositional constraints through a spatiotemporal mixture attention module and extracts fine-grained features using sparsely-activated mixture-of-experts networks. STPP~\cite{yuan2023spatio} adopts a diffusion model framework to learn complex spatiotemporal joint distributions via multi-step Gaussian diffusion and adaptively captures event dependencies with a spatiotemporal co-attention module. STFDSGCN~\cite{chang2025stfdsgcn} effectively captures the multivariate heterogeneous characteristics and dynamic spatial structures of traffic flow through dynamic sparse graph convolutional gated units and a spatiotemporal attention fusion mechanism.

Notably, Next-Generation Air Pollution Forecasting~\cite{rahmani2024next} systematically tackles three core challenges in air pollution prediction through its development of three specialized components: the PMForecast model for temporal dependencies, the GT-LSTM model for spatial dispersion mechanisms, and the FedAirNet framework for mobile sensor data privacy protection.
On the data front, the AirDelhi dataset~\cite{chauhan2023airdelhi} expands monitoring coverage through mobile sensing technology; however, since sensors are deployed exclusively on bus routes, its spatial coverage remains constrained by the bus network topology, resulting in inherent blind spots.

% Nevertheless, existing data-driven methods generally lack deep integration with physical principles, which may lead to incomplete or inaccurate representations of pollution dispersion spatiotemporal dynamics and fail to adequately reflect real physical processes. This limitation has prompted researchers to explore new approaches that combine physical constraints with data-driven models.

% Nevertheless, existing data-driven methods perform well in scenarios with sufficient and complete data, they face significant limitations in real-world air pollution prediction. The sparse distribution of monitoring stations and uneven coverage of mobile nodes often lead to incomplete and irregular spatiotemporal data, which severely hinders the accurate capture of pollution dispersion patterns. Furthermore, these methods generally lack deep integration with physical principles, limiting their ability to accurately characterize the spatiotemporal dynamics of pollution diffusion and maintain physical fidelity.

However, while existing data-driven methods demonstrate strong performance when sufficient and complete data are available, they still encounter some practical challenges in real-world air pollution forecasting. The sparse distribution of monitoring stations and uneven coverage of mobile nodes can sometimes affect the completeness and regularity of spatiotemporal data, which influences the characterization of pollution dispersion patterns. Additionally, these approaches can be enhanced through integration with physical mechanisms, as further developments in this area could help improve the representation of spatiotemporal dynamics in pollution diffusion and enhance physical consistency.

% However, these data-driven methods lack integration with physical principles, which can lead to incomplete or inaccurate representations of the spatiotemporal dynamics of pollution dispersion.

\subsubsection{Hybrid Physics-Data Driven Methods}
Physics-based and data-driven models each have advantages but integrating them remains challenging \cite{luo2025smartspr}.
HMSS \cite{Chen2020} combines PDE-based modeling for sparse areas with data-driven models for unmodeled factors, adaptively balancing both based on data and uncertainty.
AirPhyNet \cite{hettige2024airphynet} models pollutant diffusion and advection with Neural ODEs, integrating physics via graph-based latent representations.
Phy-APMR \cite{shi2025phy} integrates an inference network with a physics-informed network encoding pollutant transport PDEs, jointly minimizing data errors in observed regions and physical inconsistencies in unobserved areas.
Air-DualODE \cite{tian2024air} uses a dual-branch Neural ODE: one for physical simulation of diffusion-advection, another for spatiotemporal dependencies of environmental factors, fused to improve accuracy.
PIML~\cite{kashinath2021physics} embeds physical laws as constraints within machine learning frameworks through loss functions, network architectures, or training data to achieve integration of physical principles with data-driven methods.

% However, the nonlinear abstraction of physical variables in latent neural spaces often fails to strictly adhere to explicit physical laws, leading to systematic deviations between the generated results and physical principles. To fundamentally address this mechanistic mismatch, we propose STeP-Diff, which embeds partial differential equation constraints into the denoising process of diffusion models. This establishes an explicit connection between data fitting and physical consistency, thereby ensuring that the predictions not only align with observed data but also satisfy the fundamental requirements of physical laws.
% While existing hybrid methods attempt to incorporate physical mechanisms, the nonlinear abstraction of physical variables in their latent neural spaces often deviates from genuine physical laws. Furthermore, the complexity and large-scale nature of air pollution diffusion processes, combined with frequently incomplete domain knowledge, make accurate modeling particularly challenging.
While existing hybrid methods attempt to incorporate physical mechanisms, the nonlinear abstraction of physical variables in their latent neural spaces may not fully adhere to the underlying physical principles. Furthermore, the inherent complexity and large-scale nature of air pollution diffusion processes, combined with currently incomplete domain knowledge, present considerable challenges for precise modeling. 

% However, latent neural spaces often abstract physical variables nonlinearly, causing mismatch with explicit physics. To address this, we propose STeP-Diff, which enforces PDE constraints in denoising to ensure physically consistent predictions aligned with observed data.

% }

\subsection{Urban Spatio-Temporal Forecasting}
\label{5.2}

Urban spatio-temporal forecasting aims to model and predict the dynamic patterns of urban activities in both space and time \cite{zhao2025urbanvideo}. In recent years, deep learning technologies have achieved significant progress in this domain \cite{10679926}.
To capture spatio-temporal patterns, numerous models have been proposed, including Convolutional Neural Networks (CNN) \cite{10.1109/TKDE.2021.3135621}, Graph Neural Networks (GNN) \cite{10.1109/TKDE.2023.3333824}, Transformer \cite{10.1109/TKDE.2023.3269771},
and diffusion models \cite{yuan2023spatio}.
For instance, DGCN~\cite{gao2022novel} combines graph convolutional networks with temporal modeling to jointly enhance topological and temporal feature extraction in dynamic graphs; SDMG~\cite{zhusdmg} improves graph representation learning from a spectral perspective, enabling diffusion models to focus on low-frequency global structural information; and FGSSI~\cite{hou2025fgssi} integrates graph structure with sequence prediction to achieve efficient state inference and strong generalization in temporal propagation networks.

Furthermore, UniST has designed a universal large-scale language model that integrates various types of spatio-temporal data to efficiently forecast urban dynamics across multiple scenarios \cite{Yuan2024}. GinAR proposed a Graph Interpolation Attention Recurrent Network that accurately captures spatio-temporal dependencies in the data for efficient forecasting \cite{Yu2024}. DeepONet leverages deep neural networks to learn complex spatio-temporal relationships from distributed data streams, enabling precise predictions at specific locations at any given time \cite{Lu2021}. Meanwhile, CSDI employs a conditional score-based diffusion model to learn the correlations among different observations, thereby achieving high-precision probabilistic forecasting of time series \cite{Tashiro2021}.

% Although these models generally provide accurate predictions when sufficient data is available, in our air pollution forecasting task, the limited number of monitoring stations and the inherent randomness of mobile devices result in measurement data that are highly incomplete and temporally variable. Consequently, these factors significantly impact the forecasting performance, underscoring the need for more efficient modeling approaches.
% Although existing models generally achieve satisfactory predictive performance under conditions of high-quality data and sufficient observations, the task of air pollution forecasting faces unique challenges: the number of monitoring stations is limited, and the measurements collected by mobile monitoring devices exhibit high randomness, resulting in severe temporal sparsity and instability. Moreover, pollutant diffusion is inherently governed by physical laws, yet most current models primarily rely on purely data-driven spatio-temporal learning and fail to effectively incorporate environmental physical mechanisms such as diffusion and advection, which limits both the reliability and interpretability of the predictions. These factors collectively constrain model performance in real-world dynamic environments and underscore the necessity and urgency of developing spatio-temporal modeling approaches with enhanced physics-awareness for scenarios characterized by data sparsity and temporal evolution.
Although existing models generally achieve satisfactory predictive performance when provided with high-quality data and sufficient observations, air pollution forecasting still confronts several distinct challenges. The limited number of monitoring stations and the high randomness in data collected by mobile monitoring devices lead to discernible temporal sparsity and instability. Furthermore, although pollutant diffusion is inherently governed by physical laws, most current models, which primarily rely on purely data-driven spatiotemporal learning, exhibit certain limitations in effectively incorporating environmental physical mechanisms such as diffusion and advection. This in turn constrains the reliability and interpretability of predictions. Collectively, these factors restrict model performance in real-world dynamic environments and underscore the need to develop spatiotemporal modeling approaches with enhanced physical awareness for scenarios marked by data sparsity and temporal evolution.
\section{Conclusion and future work}
\label{con}

% In this paper, we introduce STeP-Diff for fine-grained air pollution forecasting on mobile devices. 
% By cleverly embedding DeepONet as conditional inputs into the PDE-informed diffusion model, the training process effectively captures the spatio-temporal characteristics while gradually adhering to the fundamental physical principles governing pollution diffusion, thereby ensuring the accuracy and reliability of the predictions.
% To access the performance of system, we deployed 59 self-designed portable sensing devices in the cities of Changshu and Nanjing for a 14-day period, collecting urban-scale air pollution data. 
% Compared to the second-best performing algorithm, our model achieved improvements of up to 89.12\%, 82.30\%, and 25.00\% in MAE, RMSE, and MAPE, respectively. Extensive evaluation results demonstrate that our model significantly outperforms state-of-the-art methods in air pollution prediction.
We present STeP-Diff for fine-grained air pollution forecasting on mobile devices, integrating DeepONet with a PDE-informed diffusion model to jointly capture spatio-temporal patterns and physical dynamics. Deployed across 59 sensors in Changshu and Nanjing over 14 days, STeP-Diff achieved up to 89.12\%, 82.30\%, and 25.00\% improvements in MAE, RMSE, and MAPE over the second-best method, outperforming state-of-the-art baselines.
Although the proposed model performs well in mobile air pollution forecasting, it still has limitations. Averaging sensor data within each grid may overlook local spatial heterogeneity, limiting the capture of fine-grained pollution patterns. 
Future work will explore GNN-based data fusion and promote model deployment in complex environments \cite{jian2023path, wang2024toward} and on UAV platforms\cite{wang2024transformloc}.

% }

\bibliographystyle{IEEEtran}
\bibliography{infocom}

\newpage
\appendices

% {\color{blue}
\section{Proof of Theorm 1}
\label{proof}
\setcounter{theorem}{0}
\begin{theorem}
Define the noise term as:
\begin{equation}
\epsilon^{l+1}_t=G(\epsilon^l_t)=B\epsilon_t^l+C_tS^l,
\tag{12}
\label{noise_PDE}
\end{equation}
where $G(\cdot)$ is the PDE evolution operator defined in Eq. (\ref{PDE-l}). Then, the pollutant concentration satisfies:
\begin{equation}
{V_0}^{l+1}=G({V_0}^l).
\tag{13}
\label{pollution_PDE}
\end{equation}
\end{theorem}

\begin{proof}

First, according to the noise-added equation (Eq. (\ref{noise-added})) for pollutant concentration, we obtain:
\begin{equation}
    V_0^l = \frac{V_T^l - (1-\alpha_T)\epsilon^l}{\sqrt{\alpha_T}},
\label{l}
\end{equation}
\begin{equation}
    V_0^{l+1} = \frac{V_T^{l+1}  - (1-\alpha_T)\epsilon^{l+1}}{\sqrt{\alpha_T}}.
\label{l+1}
\end{equation}
Next, based on Eq. (\ref{noise-added}) and Eq. (\ref{noise_PDE}), we can derive:
\begin{equation}
\begin{aligned}
    \epsilon^{l+1} &=\sum_{j=1}^T\epsilon_j^{l+1}=\sum_{j=1}^T G(\epsilon^l_j)=
    G\left(\sum_{j=1}^T \epsilon^l_j\right) \\
    &=G(\epsilon^{l})=B\epsilon^l+C_tS^l.
\end{aligned}
\label{noise_v2}
\end{equation}
Substituting Eq. (\ref{noise_v2}) into Eq. (\ref{l+1}), we can rewrite $V_0^{l+1}$ as:
\begin{equation}
    V_0^{l+1}=\frac{V_T^{l+1}}{\sqrt{\alpha_T}}-\frac{1-\alpha_T}{\sqrt{\alpha_T}}\left(B \epsilon^l + C S^l\right).
\label{re_V}
\end{equation}
Next, applying the PDE operator $G(\cdot)$ to $V_0^{l}$, and combining Eq. (\ref{PDE-l}) and Eq. (\ref{l}), we obtain:
\begin{equation}
\begin{aligned}
    G(V_0^l) &=B \left( \frac{V_T^l - (1-\alpha_T)\epsilon^l}{\sqrt{\alpha_T}} \right)+ C S^l \\
    &= B \frac{V_T^l}{\sqrt{\alpha_T}}-B \frac{1-\alpha_T}{\sqrt{\alpha_T}} \epsilon^l+ C S^l.
\end{aligned}
\label{G_V}
\end{equation}
Then, calculating the difference between $V_0^{l+1}$ and $G(V_0^l)$, and combining Eq. (\ref{re_V}) and Eq. (\ref{G_V}), we get:
\begin{equation}
\begin{aligned}
    V_0^{l+1} - G(V_0^l) &=\frac{V_T^{l+1}}{\sqrt{\alpha_T}}-\frac{1-\alpha_T}{\sqrt{\alpha_T}} B \epsilon^l-\frac{1-\alpha_T}{\sqrt{\alpha_T}} C S^l \\
    &- \frac{V_T^l}{\sqrt{\alpha_T}} B+ \frac{1-\alpha_T}{\sqrt{\alpha_T}} B \epsilon^l- C S^l \\
    &=\frac{V_T^{l+1}}{\sqrt{\alpha_T}}- \frac{V_T^l}{\sqrt{\alpha_T}} B-\frac{1-\alpha_T}{\sqrt{\alpha_T}} C S^l- C S^l .
\end{aligned}
\label{V_diff}
\end{equation}
\textit{The first two terms} are linear transformations of $V_T^{l+1}$ and $V_T^{l}$, respectively. Since both are independent zero-mean Gaussian variables, their difference also follows a zero-mean Gaussian distribution, which can be regarded as random perturbation.
\textit{The last two terms} are deterministic bias terms. When $S^l=0$ (i.e., no pollution source), this bias is zero. When $S^l\neq 0$, the bias is fixed. In practical model training, such biases can be compensated by the learnable parameters, and thus do not affect the physical consistency of the model.

Therefore, Eq. (\ref{V_diff}) can be written as:
\begin{equation}
\begin{aligned}
    V_0^{l+1} &= G(V_0^l) 
    +\underbrace{\text{zero-mean Gaussian noise}}_{\text{due to independent sampling}} \\
    & + \underbrace{\text{deterministic bias terms}}_{\text{compensable by model}} \\
    & \approx G(V_0^l),
\end{aligned}
\end{equation}
which proves the theorem.

\textbf{Remark.} Due to the sampling process in diffusion models, zero-mean Gaussian noise is introduced, which can be interpreted as unmodeled random perturbations. The deterministic bias can be explicitly compensated, ensuring that the learned operator $G(\cdot)$ still faithfully captures the physical spatio-temporal evolution of pollutant concentration. Moreover, in our experimental setup, there is no pollution source term, i.e., $S^l=0$.

\end{proof}

\section{Statistical Data Analysis}
\label{A2}
% \subsection{Statistical Data Analysis}
% \label{Sta}

\begin{figure}[t!]
  \centering
  \subfigure[Nanjing.]{
    \centering
    \includegraphics[height=0.45\linewidth]{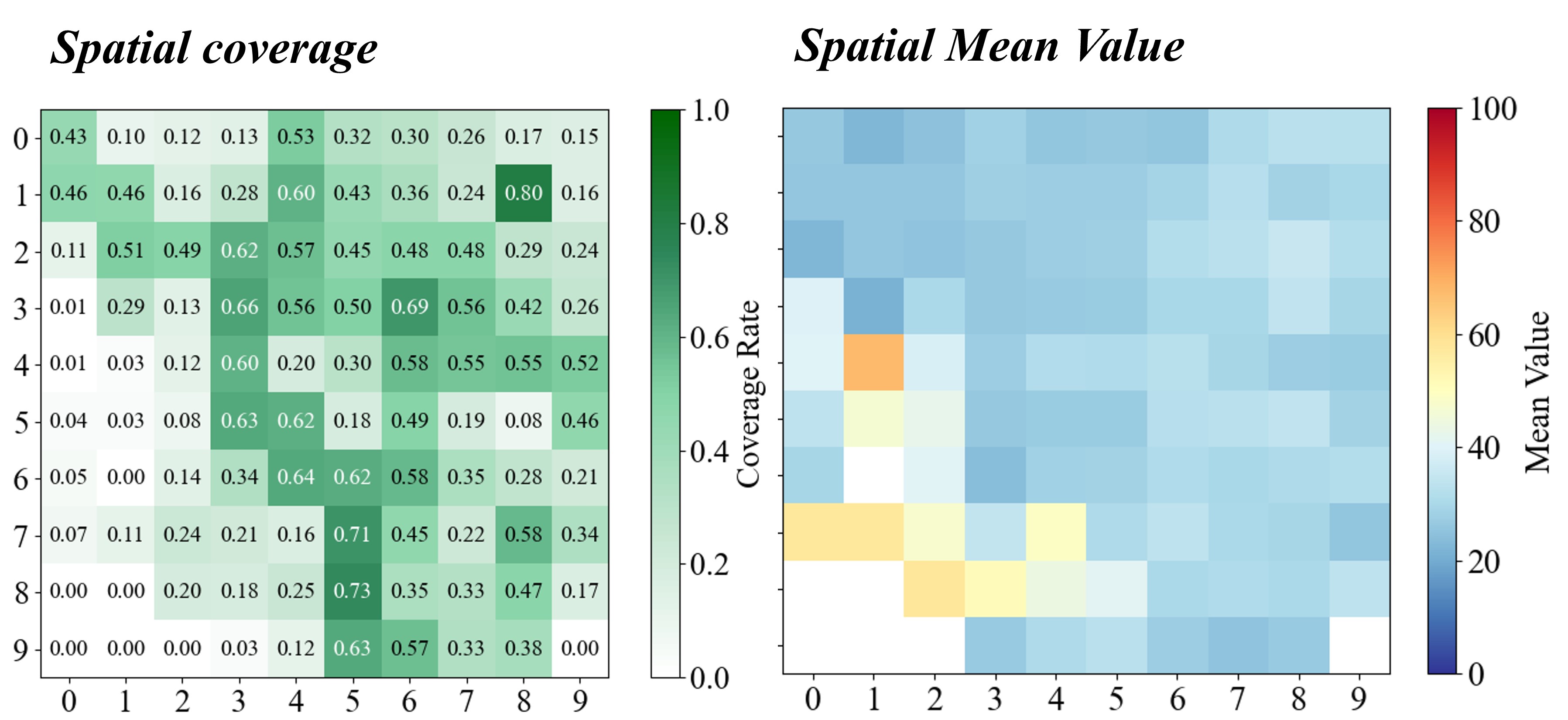}
  }
  \hfill
  \subfigure[Changshu.]{
    \centering
    \includegraphics[height=0.45\linewidth]{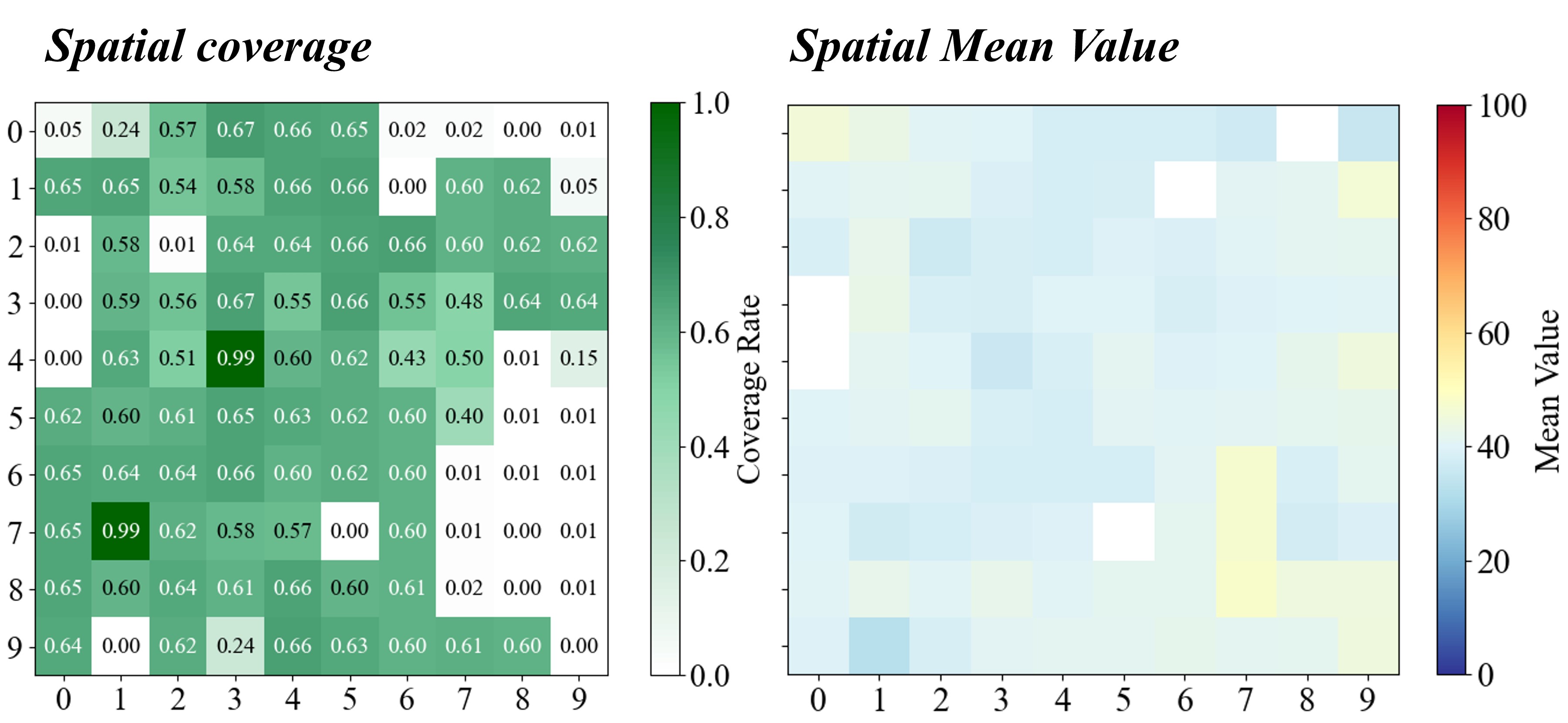}
  }
  \caption{Spatial coverage per time slice and average PM$_{2.5}$ measurement for each time slice within the selected time period.}
  \label{fig:time}
\end{figure}

To assess data coverage in real-world scenarios and its temporal variations, Fig. \ref{fig:time} presents the hourly spatial data coverage and PM$_{2.5}$ concentration measurements over a 14-day period. The upper part of the figure reveals that data coverage in Nanjing, collected from cars, exhibits significant fluctuations, whereas in Changshu, where buses operate along fixed routes and schedules, coverage follows a more regular and periodic pattern. Both cities experience the lowest coverage during the early morning hours, approaching zero.
The lower part of the figure illustrates the temporal variations in PM$_{2.5}$ concentrations. The horizontal lines represent the average concentration within each time interval, while the vertical lines indicate the range, with the highest and lowest points corresponding to the maximum and minimum recorded concentrations, respectively. Notably, Nanjing exhibits greater fluctuations in measurements within each time interval, whereas in Changshu, the observed values remain closer to the mean. This discrepancy is largely influenced by urban activity patterns and the operational characteristics of mobile sensing devices in the selected subregions.
Despite the large number of samples collected in each hourly time slice, the sensed subregions remain notably incomplete, highlighting the challenges of achieving comprehensive spatio-temporal coverage in mobile sensing-based air pollution monitoring.

\begin{figure}[t!]
  \centering
  \subfigure[Nanjing.]{
    \centering
    \includegraphics[height=0.6\linewidth]{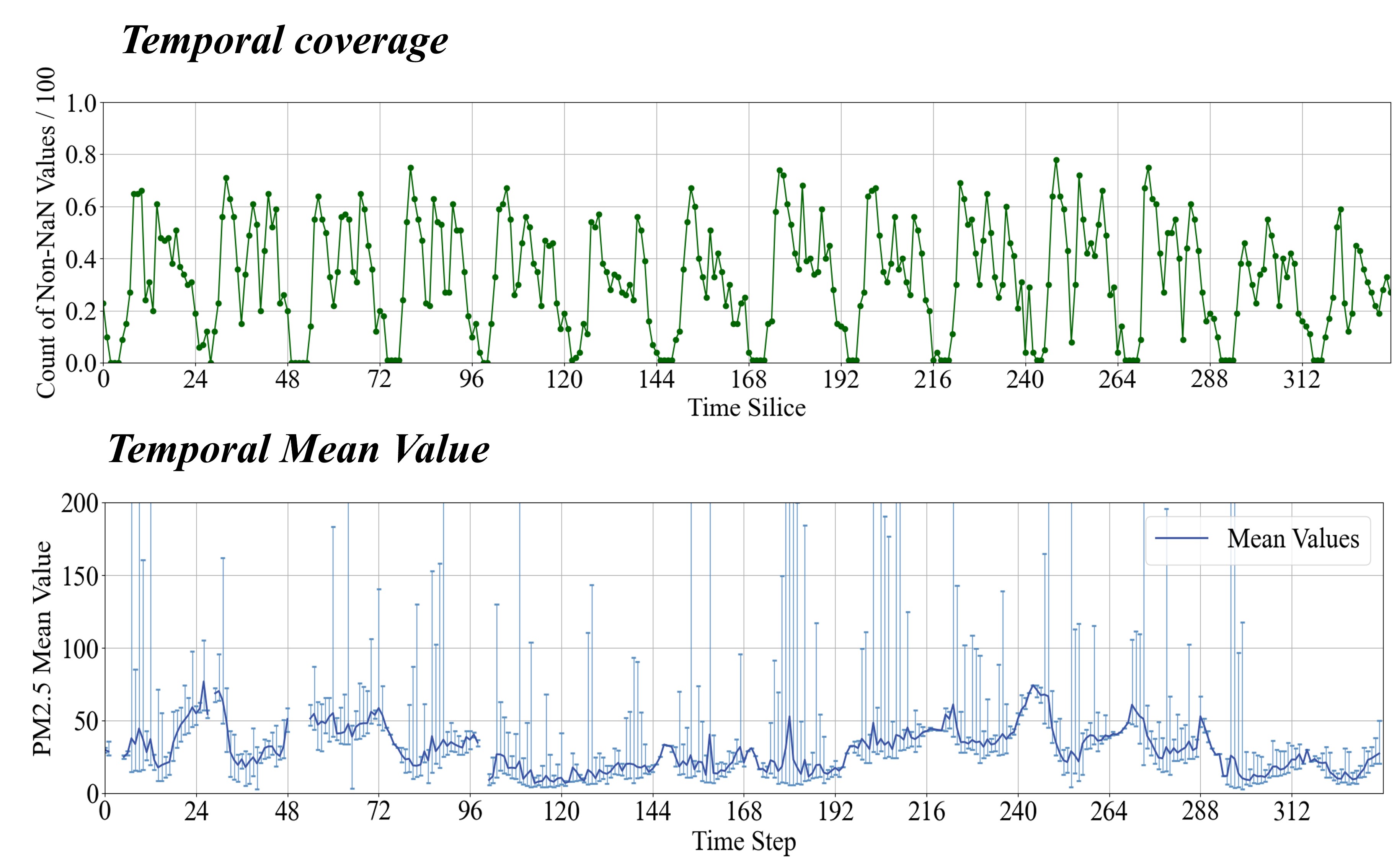}
  }
  \hfill
  \subfigure[Changshu.]{
    \centering
    \includegraphics[height=0.6\linewidth]{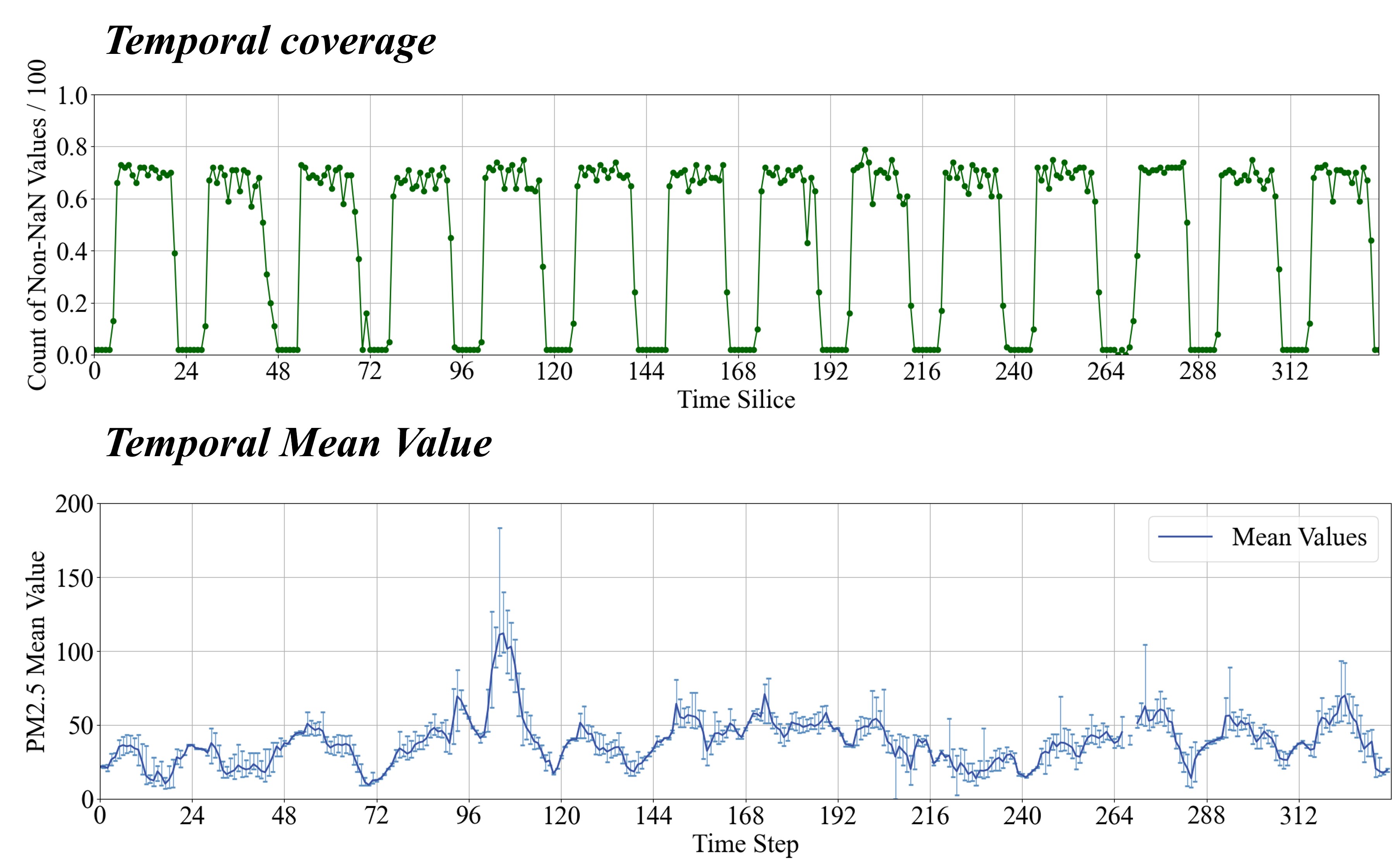}
  }
  \caption{Time coverage per grid and average PM$_{2.5}$ measurement within each grid in the selected sub-region across the total time frame.}
  \label{fig:space}
\end{figure}

To further analyze the data, the left part of Fig. \ref{fig:space}(a)(b) shows the temporal coverage rate for each grid over the entire time span. As the coverage rate increases, the green color deepens. It can be observed that in Nanjing, the temporal coverage varies significantly across different locations due to the free movement of cars, while in Changshu, the coverage provided by public buses, operating along fixed routes, is relatively uniform across locations. However, both cities exhibit regions with no observational data at all, resulting in a coverage rate of zero for those areas.
The right part of Fig. \ref{fig:space}(a)(b) presents the average PM$_{2.5}$ concentration for each grid across all time intervals. As the concentration increases, the color blocks transition from blue to yellow and then to red. In Nanjing, the average concentrations vary greatly between regions, with some areas having an average concentration as low as 30 $\mu g/m^3$, while others reach as high as 50 $\mu g/m^3$; in contrast, the average concentration in Changshu is generally close to 50 $\mu g/m^3$. These findings highlight two major challenges: incomplete and time-varying data coverage, which significantly complicate the task of obtaining accurate fine-grained air pollution estimates.

\begin{table*}
\caption{Sensor specifications of the portable sensing devices.}
\label{sensor}
\centering
\normalsize
% \fontsize{8pt}{12pt}\selectfont % 手动设置字号（12pt 字号，14pt 行距）
% \setlength{\tabcolsep}{3pt} % 设置列间距
\begin{tabular}{cccccc}
\toprule
\textbf{Sensor} & \textbf{Gas Type} & \textbf{Noise} & \textbf{Range} & \textbf{Sensitivity} & \textbf{Tested T$_{90}$} \\
\midrule
PMS5003T & PM$_{2.5}$/PM$_{10}$ & $\begin{array}{l} \pm 10 \mu g/m^3 @ <100 \mu g/m^3 \\ \pm 10\% @ >100 \mu g/m^3 \end{array}$ & 0-1000$\mu g/m^3$ & $\begin{array}{l}  50\%-0.3\mu m \\ 98\%-0.5\mu m  \end{array}$  & $\geq$60s \\
OX-B431 & O$_3$ & $\pm 15$ppb & 0-20ppm & -225$\sim$-550nA/ppm & $\geq$ 60s \\
SO2-B4 & SO$_2$ & $\pm 5$ppb & 0-100ppm & 275$\sim$475nA/ppm & $\geq$ 80s \\
NO2-B43F & NO$_2$ & $\pm 15$ppb & 0-20ppm & -175$\sim$-450nA/ppm  & $\geq$ 40s \\
CiTiceL@4CM  & CO & $\pm 4$ppb & 0-2000ppm & 70$\pm$15nA/ppm & $\geq$ 30s \\
\bottomrule
\end{tabular}
\end{table*}

% \begin{figure*}[ht]
%   \centering
%   \subfigure[Nanjing.]{
%     \centering
%     \includegraphics[height=3.45 cm]{samples/Fig/NJ Trace.jpg}
%   }
%   \hfill
%   \subfigure[Changshu.]{
%     \centering
%     \includegraphics[height=3.45 cm]{samples/Fig/CS Trace.jpg}
%   }
%   \caption{The operational trajectory of the mobile devices and the locations of fixed monitoring stations. The lines in different colors represent the trajectories of different vehicles, while the black triangles indicate the locations of the fixed monitoring stations.}
%   \label{fig:trace}
% \end{figure*}

\section{Sliding Window Method for Dataset Partitioning}
\label{A1}

\begin{figure}[t!]
  \centering
  \includegraphics[width=1\linewidth]{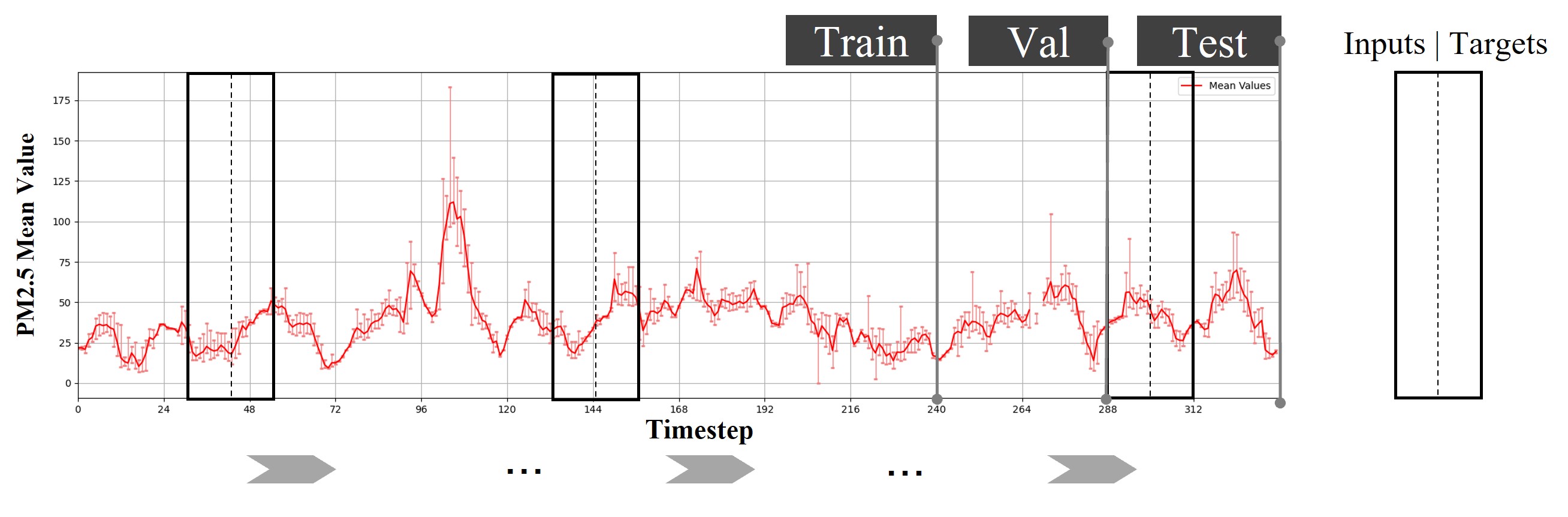}
  \caption{Taking the PM$_{2.5}$ concentration measurements from Changshu as an example, the dataset is divided using a sliding window. The red line in the figure represents the variation in the average value of each time slice over time. Each sliding window includes 12 time steps as input and 12 time steps as the target prediction values.}
  \label{fig:slide}
\end{figure}

As shown in Fig. \ref{fig:slide}, the original spatio-temporal field is divided along the time dimension (the first dimension) into training, validation, and test sets with a 5:1:1 ratio. A sliding window of size "Inputs + Targets" is used to generate the samples. All experiments use the same setting of "12+12," which means utilizing the past 12 time steps to predict the subsequent 12 time steps. Therefore, the measurement data for each selected spatio-temporal region in the city is divided into training sets of size $\{217, 24, 10, 10\}$, validation sets of size $\{25, 24, 10, 10\}$, and test sets of size $\{25, 24, 10, 10\}$, ensuring no data leakage. 
% We train STeP-Diff on the training and validation sets and evaluate the model's performance on the test set.

\section{Physics-Informed Combinations and Performance}
\label{A3}

\begin{table*}[t]
\caption{
% \color{blue}
P-values from t-tests comparing our model to the second-best baseline
}
\label{p-value}
\centering
\normalsize
\renewcommand{\arraystretch}{0.95}
\begin{tabularx}{\textwidth}{l@{\hskip 1pt}*{12}{>{\centering\arraybackslash}p{1.4cm}@{\hskip 0.0pt}}}
\toprule
\textbf{City} & \multicolumn{3}{c}{\textbf{Changshu (Mobile)}} & \multicolumn{3}{c}{\textbf{Nanjing (Mobile)}} & \multicolumn{3}{c}{\textbf{Changshu (National)}} & \multicolumn{3}{c}{\textbf{Nanjing (National)}} \\
\cmidrule(lr){2-4} \cmidrule(lr){5-7} \cmidrule(lr){8-10} \cmidrule(lr){11-13}
\textbf{Metrics} & MAE & RMSE & MAPE & MAE & RMSE & MAPE & MAE & RMSE & MAPE & MAE & RMSE & MAPE \\
\midrule
p-value & $8.95\times10^{-8}$ & $1.23 \times 10^{-7}$ & $2.00 \times 10^{-2}$ & $3.72 \times 10^{-6}$ & $1.35 \times 10^{-5}$ & $2.85 \times 10^{-2}$ & $9.08 \times 10^{-4}$ & $4.34 \times 10^{-3}$ & $2.10 \times 10^{-2}$ & $1.42 \times 10^{-5}$ & $1.21 \times 10^{-4}$ & $5.34 \times 10^{-2}$ \\
\bottomrule
\end{tabularx}
\end{table*}

\begin{table*}[t]
\caption{Performance of physics-informed combinations experiments, where bold denotes the best results and underline denotes the optimal combination of core components.}
\label{combine}
\centering
\normalsize
\setlength{\tabcolsep}{5pt} % 调整列间距，可以根据需要调整
\begin{tabularx}{\textwidth}{l*{12}{>{\centering\arraybackslash}p{0.8cm}}}
\toprule
\textbf{City} & \multicolumn{3}{c}{\textbf{Changshu (Mobile)}} & \multicolumn{3}{c}{\textbf{Nanjing (Mobile)}} & \multicolumn{3}{c}{\textbf{Changshu (National)}} & \multicolumn{3}{c}{\textbf{Nanjing (National)}} \\
\cmidrule(lr){2-4} \cmidrule(lr){5-7} \cmidrule(lr){8-10} \cmidrule(lr){11-13}
\textbf{Metrics} & MAE & RMSE & MAPE & MAE & RMSE & MAPE & MAE & RMSE & MAPE & MAE & RMSE & MAPE \\
\midrule
\textbf{Diff} & 8.09 & 16.47 & 1.60 & 3.34 & 8.80 & 0.56 & 8.95 & 18.45 & 0.99 & 2.50 & 7.44 & 0.33 \\
\midrule
% {\color{blue}\textbf{DeepONet}} & \textcolor{blue}{10.41} & \textcolor{blue}{13.47} & \textcolor{blue}{0.30} & \textcolor{blue}{15.69} & \textcolor{blue}{24.27} & \textcolor{blue}{0.52} & \textcolor{blue}{8.97} & \textcolor{blue}{12.85} & \textcolor{blue}{0.28} & \textcolor{blue}{11.97} & \textcolor{blue}{14.51} & \textcolor{blue}{0.77} \\
\textbf{DeepONet} & 10.41 & 13.47 & 0.30 & 15.69 & 24.27 & 0.52 & 8.97 & 12.85 & 0.28 & 11.97 & 14.51 & 0.77 \\
\midrule
\textbf{Diff + DeepONet} & & & & & & & & & & & \\
\textcircled{\scriptsize 1} Frozen & 7.10 & \underline{13.78} & \underline{0.37} & 2.57 & 7.38 & 0.57 & 6.99 & \underline{13.50} & \underline{0.32} & 2.48 & 6.83 & 0.58 \\
\textcircled{\scriptsize 2} Trainable & \underline{6.74} & 17.81 & 0.72 & \underline{1.30} & \underline{4.01} & \underline{0.55} & \underline{6.64} & 17.58 & 0.57 & \underline{1.29} & \underline{3.97} & \underline{0.35} \\
\midrule
\textbf{Diff + PDE} & & & & & & & & & & & \\
\textcircled{\scriptsize 3} 1-city fit & 7.17 & 17.95 & 0.76 & 9.28 & 19.70 & 0.76 & 7.36 & 18.85 & 0.88 & 8.87 & 19.28 & 0.59 \\
\textcircled{\scriptsize 4} 4-city fit & 2.42 & \underline{9.17} & \underline{0.26} & 1.99 & 6.06 & \underline{0.22} & \underline{2.40} & \underline{7.00} & \underline{0.25} & 1.91 & 5.89 & \underline{0.47} \\
\textcircled{\scriptsize 5} Diff\_loss & \underline{2.25} & 11.56 & 1.18 & \underline{1.64} & \underline{4.61} & 0.43 & 3.50 & 12.90 & 0.93 & \underline{1.38} & \underline{3.88} & 0.73 \\
\midrule
\textbf{Diff + PDE + DeepONet} & & & & & & & & & & & \\
\textcircled{\scriptsize 6} 1-city fit & 4.96 & 12.85 & 0.65 & 8.60 & 21.52 & 1.12 & 4.98 & 12.73 & 0.69 & 8.44 & 21.31 & 1.32 \\
\textcircled{\scriptsize 7} 4-city fit & 22.39 & 39.60 & 1.49 & 3.49 & 9.17 & 0.56 & 22.09 & 39.60 & 1.52 & 3.40 & 8.61 & 0.60 \\
\textcircled{\scriptsize 8} random noise fit & 15.74 & 34.18 & 0.40 & 1.52 & 3.99 & 0.59 & 15.92 & 34.37 & 0.39 & 1.57 & 4.21 & \underline{\textbf{0.27}} \\
\textcircled{\scriptsize 9} DeepONet\_loss & 32.02 & 65.45 & 0.35 & 28.67 & 44.50 & 1.43 & 31.68 & 64.98 & 0.37 & 28.52 & 44.17 & 1.66 \\
\textcircled{\scriptsize 10} Diff\_loss & \underline{\textbf{1.30}} & \underline{\textbf{3.52}}& \underline{\textbf{0.17}} & \underline{\textbf{0.95}} & \underline{\textbf{2.94}} & \underline{\textbf{0.15}} & \underline{\textbf{0.88}} & \underline{\textbf{2.40}} & \underline{\textbf{0.25}} & \underline{\textbf{0.84}} & \underline{\textbf{3.25}} & 0.41 \\
\bottomrule
\end{tabularx}
\end{table*}

Table \ref{combine} reports the results of our studies. For clarity, we have abbreviated the descriptive labels in the first column. “Diff” denotes the baseline model employing only the diffusion module \cite{Tashiro2021}. The “Diff + DeepONet” group distinguishes two configurations: \textcircled{\scriptsize 1} the diffusion module is pre-trained, and the information learned by DeepONet is only added as conditional input during the prediction phase, and \textcircled{\scriptsize 2} the diffusion module is trainable with DeepONet used as a conditional input. 
Similarly, the "Diff + PDE" group compares three settings: \textcircled{\scriptsize 3} and \textcircled{\scriptsize 4} correspond to models where the PDE parameters are fitted using data from a single city or four cities, respectively, with the PDE predictions used as conditional input to the diffusion model; whereas \textcircled{\scriptsize 5} represents the incorporation of the PDE equation as a physical constraint directly into the loss function of the diffusion model. 
Finally, the "Diff + PDE + DeepONet" group explores different combinations of the three components, with the primary distinction in how the PDE is utilized. Configurations \textcircled{\scriptsize 6} \textcircled{\scriptsize 7} \textcircled{\scriptsize 8} fit the PDE using measurements from a single city, four cities, and random noise, respectively, and use both the PDE and DeepONet predictions as conditional inputs to the diffusion model. Configurations \textcircled{\scriptsize 9} and \textcircled{\scriptsize 10} incorporate the PDE equation as a physical constraint into the loss function of DeepONet and the diffusion model, respectively, and use the DeepONet predictions as conditional input to the diffusion model.

% It is evident that configuration \textcircled{\scriptsize 10}, which incorporates the PDE equation as a physical constraint within the loss function of the diffusion model while using the spatial information learned by DeepONet as a conditional input, delivers the best performance. 
% {\color{blue}
The results clearly demonstrate that configuration \textcircled{\scriptsize 10} outperforms all other settings, including those using only the diffusion model, only DeepONet, or their various combinations. This optimal performance stems from the integration of the PDE equation as a physical constraint within the loss function of the diffusion model, together with the spatial information learned by DeepONet used as conditional input, enabling a more accurate and physically consistent prediction.
% }
When the PDE equation is used as a physical constraint in the loss function of the diffusion model (configuration \textcircled{\scriptsize 5}), compared to the baseline diffusion model (configuration \textcircled{\scriptsize 1}), significant improvements in MAE, RMSE, and MAPE are observed across the four datasets: 72.19\%, 29.81\%, and 26.25\%; 50.90\%, 47.61\%, and 23.21\%; 60.90\%, 30.08\%, and 6.06\%; and 44.80\%, 47.85\%, and -121.21\%, respectively.Building on configuration \textcircled{\scriptsize 5}, incorporating the spatial information learned by DeepONet as a conditional input (configuration \textcircled{\scriptsize 10}) further boosts the model's performance, yielding additional improvements in MAE, RMSE, and MAPE of 42.22\%, 69.55\%, and 85.60\%; 42.07\%, 79.61\%, and 65.12\%; 74.86\%, 81.40\%, and 73.12\%; and 9.13\%, 16.24\%, and 43.8\% across the four datasets, respectively. These results clearly demonstrate that the optimal performance enhancement is achieved through the integration of the diffusion model, PDE, and DeepONet.

% {\appendix[Proof of the Zonklar Equations]
% Use $\backslash${\tt{appendix}} if you have a single appendix:
% Do not use $\backslash${\tt{section}} anymore after $\backslash${\tt{appendix}}, only $\backslash${\tt{section*}}.
% If you have multiple appendixes use $\backslash${\tt{appendices}} then use $\backslash${\tt{section}} to start each appendix.
% You must declare a $\backslash${\tt{section}} before using any $\backslash${\tt{subsection}} or using $\backslash${\tt{label}} ($\backslash${\tt{appendices}} by itself
%  starts a section numbered zero.)}

% {
% \color{blue}
\section{Training and Inference Time Comparison of All Methods}
\label{time}

\begin{table*}[t]
\caption{Training and Inference Time Comparison of All Methods.}
\label{Table Time}
\centering
\normalsize
\begin{tabularx}{\textwidth}{l@{\hskip 1pt}*{9}{>{\centering\arraybackslash}p{1.95cm}@{\hskip 0.05pt}}}
\toprule
\textbf{City} & \multicolumn{2}{c}{\textbf{Changshu (Mobile)}} & \multicolumn{2}{c}{\textbf{Nanjing (Mobile)}} & \multicolumn{2}{c}{\textbf{Changshu (National)}} & \multicolumn{2}{c}{\textbf{Nanjing (National)}} \\
\cmidrule(lr){2-3} \cmidrule(lr){4-5} \cmidrule(lr){6-7} \cmidrule(lr){8-9}
\textbf{Metrics} & Training Time (second) & Inference Time (second) & Training Time (second) & Inference Time (second) & Training Time (second) & Inference Time (second) & Training Time (second) & Inference Time (second) \\
\midrule
PDE & 0.03 & 0.002 & 0.03 & 0.001 & 0.03 & 0.001 & 0.03 & 0.001 \\
\midrule
NeuralODE & 166.26 & 0.57 & 199.43 & 0.49 & 191.04 & 0.71 & 177.63 & 0.41 \\
GinAR & 227.86 & 1.18 & 241.76 & 1.13 & 252.51 & 1.23 & 263.34 & 1.12 \\
UniST & 108.02 & 2.11 & 106.04 & 2.41 & 70.86 & 2.13 & 72.84 & 2.25 \\
DDPM & 110.51 & 42.21 & 124.62 & 24.64 & 134.98 & 31.24 & 114.65 & 32.24 \\
DeepONet & 16.26 & 0.02 & 17.24 & 0.01 & 16.68 & 0.02 & 16.74 & 0.02 \\
CSDI & 129.58 & 47.24 & 166.81 & 38.06 & 150.93 & 36.57 & 129.94 & 37.82 \\
STGNN & 17.67 & 0.18 & 17.43 & 0.13 & 16.46 & 0.17 & 17.06 & 0.14 \\
AirFormer & 11.29 & 0.26 & 11.56 & 0.23 & 7.40 & 0.16 & 11.66 & 0.12 \\
% \textcolor{blue}{STGNN} & \textcolor{blue}{17.67} & \textcolor{blue}{0.18} & \textcolor{blue}{17.43} & \textcolor{blue}{0.13} & \textcolor{blue}{16.46} & \textcolor{blue}{0.17} & \textcolor{blue}{17.06} & \textcolor{blue}{0.14} \\
% \textcolor{blue}{AirFormer} & \textcolor{blue}{11.29} & \textcolor{blue}{0.26} & \textcolor{blue}{11.56} & \textcolor{blue}{0.23} & \textcolor{blue}{7.40} & \textcolor{blue}{0.16} & \textcolor{blue}{11.66} & \textcolor{blue}{0.12} \\
\midrule
HMSS & 17.80 & 1.02 & 18.70 & 0.97 & 19.20 & 1.00 & 18.40 & 0.98 \\
AirPhyNet & 136.88 & 0.10 & 48.66 & 0.21 & 84.50 & 0.19 & 29.59 & 0.10 \\
Phy-AMPR & 162.70 & 2.47 & 181.40 & 2.39 & 176.60 & 2.41 & 168.90 & 2.54 \\
DiffusionPDE & 124.52 & 110.21 & 154.98 & 124.62 & 146.25 & 114.21 & 151.62 & 116.14 \\
PINN-Diffusion & 132.70 & 88.51 & 165.22 & 98.20 & 197.61 & 75.12 & 187.51 & 77.61 \\
% \textcolor{blue}{DiffusionPDE} & \textcolor{blue}{124.52} & \textcolor{blue}{110.21} & \textcolor{blue}{154.98} & \textcolor{blue}{124.62} & \textcolor{blue}{146.25} & \textcolor{blue}{114.21} & \textcolor{blue}{151.62} & \textcolor{blue}{116.14} \\
% \textcolor{blue}{PINN-Diffusion} & \textcolor{blue}{132.70} & \textcolor{blue}{88.51} & \textcolor{blue}{165.22} & \textcolor{blue}{98.20} & \textcolor{blue}{197.61} & \textcolor{blue}{75.12} & \textcolor{blue}{187.51} & \textcolor{blue}{77.61} \\
\midrule
STeP-Diff & 100.87 & 61.98 & 141.22 & 61.61 & 132.73 & 61.01 & 126.17 & 61.51 \\
\bottomrule
\end{tabularx}
\end{table*}

As shown in Table~\ref{Table Time}, we compare the training and inference times of several representative models on four cities and two types of ground truth using a single NVIDIA RTX A6000 GPU, and conduct a systematic analysis of the time efficiency of the proposed STeP-Diff method.

In terms of training time, STeP-Diff consistently demonstrates above-average training efficiency across different scenarios. It significantly outperforms high-overhead models such as GinAR, NeuralODE, and Phy-AMPR in terms of training speed, indicating favorable trainability. Although slightly slower than lightweight models like DeepONet and HMSS, STeP-Diff offers greater capacity and expressive power.

Regarding inference time, STeP-Diff takes approximately 61 seconds in all settings. Compared to other diffusion-based models such as DDPM and CSDI, STeP-Diff is relatively slower in inference, but this is offset by its superior predictive accuracy. When compared to non-diffusion models like DeepONet and AirPhyNet, which exhibit extremely low inference latency, STeP-Diff shows a clear disadvantage in response time. This indicates that the current version of STeP-Diff is not yet suitable for real-time applications, but is more advantageous in offline forecasting scenarios where predictive accuracy is critical.

In summary, STeP-Diff achieves a well-balanced trade-off between training efficiency and model performance, making it well-suited for high-quality forecasting tasks on medium-scale datasets. In future work, we plan to incorporate fast sampling strategies or model compression techniques to further improve inference efficiency, enhancing its applicability in real-time and deployment-sensitive environments.
\section{Decision-Oriented Evaluation under a Fixed PM$_{2.5}$ Warning Threshold}
\label{warn}

\begin{table}[t!]
\caption{Evaluation of Pollution Alert Performance for All Methods. The ground-truth labels in both cities are obtained solely from national stations.}
\label{warn}
\centering
\normalsize
\begin{tabularx}{\linewidth}{l@{\hskip 1pt}*{4}{>{\centering\arraybackslash}p{1.6cm}@{\hskip 0.3pt}}}
\toprule
\textbf{City} & \multicolumn{2}{c}{\textbf{Changshu}} & \multicolumn{2}{c}{\textbf{Nanjing}} \\
\cmidrule(lr){2-3} \cmidrule(lr){4-5}
\textbf{Metrics} & Recall & F1 Score & Recall & F1 Score \\
\midrule
PDE & 0.50 & 0.40 & 0.50 & 0.50 \\
\midrule
NeuralODE & 0.25 & 0.22 & 0.50 & 0.57 \\
GinAR & 0.75 & 0.67 & 0.75 & 0.67 \\
UniST & 0.75 & 0.75 & 0.25 & 0.25 \\
DDPM & 0.25 & 0.33 & 0.50 & 0.67 \\
DeepONet & 0.75 & 0.67 & 0.75 & 0.75 \\
CSDI & 0.75 & 0.75 & 0.50 & 0.67 \\
STGNN & 0.50 & 0.67 & 0.50 & 0.67 \\
AirFormer & 0.75 & 0.75 & 0.75 & 0.75 \\
% STGNN & \textcolor{blue}{0.50} & \textcolor{blue}{0.67} & \textcolor{blue}{0.50} & \textcolor{blue}{0.67} \\
% AirFormer & \textcolor{blue}{0.75} & \textcolor{blue}{0.75} & \textcolor{blue}{0.75} & \textcolor{blue}{0.75} \\
\midrule
HMSS & 0.50 & 0.67 & 0.75 & 0.67 \\
AirPhyNet & 0.75 & 0.67 & 0.75 & 0.75 \\
Phy-AMPR & 0.75 & 0.73 & 0.75 & 0.67 \\
DiffusionPDE & 0.50 & 0.67 & 0.50 & 0.67 \\
PINN-Diffusion & 0.75 & 0.75 & 0.75 & 0.76 \\
% DiffusionPDE & \textcolor{blue}{0.50} & \textcolor{blue}{0.67} & \textcolor{blue}{0.50} & \textcolor{blue}{0.67} \\
% PINN-Diffusion & \textcolor{blue}{0.75} & \textcolor{blue}{0.75} & \textcolor{blue}{0.75} & \textcolor{blue}{0.76} \\
\midrule
STeP-Diff & 1.00 & 0.89 & 1.00 & 0.89 \\
\bottomrule
\end{tabularx}
\end{table}

To further evaluate the practical applicability of the proposed model in real-world pollution control scenarios, we introduce a decision-oriented evaluation method based on a fixed pollution threshold. Following the World Health Organization (WHO) Air Quality Guidelines, which recommend a 24-hour average PM$_{2.5}$ concentration not exceeding 25~$\mu g/m^3$ to minimize short-term exposure risks~\cite{world2006air}, this threshold is widely recognized for its health-protective significance. Therefore, in this study, a warning threshold of 25~$\mu g/m^3$ is set, whereby predicted 24-hour PM$_{2.5}$ concentrations exceeding this value are classified as pollution days, triggering alerts to support timely mitigation measures.

Within this framework, we extend model evaluation beyond traditional regression error metrics (e.g. MAE, MSE) to classification-based metrics that better reflect decision-making effectiveness. Specifically, we adopt Recall to measure the model’s ability to identify pollution days, emphasizing the reduction of missed warnings and associated health risks, and employ F1 Score as a comprehensive metric to balance false positives and false negatives, thereby providing a holistic assessment of warning quality. The metrics are defined as follows:
\begin{align}
\text{Recall} &= \frac{TP}{TP + FN}, \\
\text{Precision} &= \frac{TP}{TP + FP}, \\
\text{F1 Score} &= 2 \times \frac{\text{Precision} \times \text{Recall}}{\text{Precision} + \text{Recall}},
\end{align}
where $TP$ denotes true positives (correctly predicted pollution days), $FN$ denotes false negatives (missed pollution days), and Precision represents the proportion of predicted pollution days that are actual pollution days.

The study forecasts PM$_{2.5}$ 24-hour averages for 10 consecutive days and evaluates model performance against national monitoring station data from two cities. As shown in Table~\ref{warn}, under the 25~$\mu g/m^3$ warning threshold, STeP-Diff demonstrates superior pollution warning capability on national monitoring data from both Changshu and Nanjing. 
% The model achieves a Recall of 1.00 in both cities, indicating no missed pollution days, and attains the highest F1 Scores of 0.89 and 1.00 in Changshu and Nanjing, respectively, significantly outperforming other methods. 
The model achieves a Recall of 1.00 in both cities, indicating no missed pollution days, and an F1 Score of 0.89 in both cases, significantly outperforming other methods.
These results further validate the robustness and practical utility of STeP-Diff in real-world pollution warning scenarios.
% }

\end{document}